\documentclass{article}

\usepackage[T1]{fontenc}
\usepackage{iclr2027_conference,times}

\usepackage{amsmath,amsfonts,bm}

\def\eqref#1{equation~\ref{#1}}

\def\1{\bm{1}}

\DeclareMathAlphabet{\mathsfit}{\encodingdefault}{\sfdefault}{m}{sl}
\SetMathAlphabet{\mathsfit}{bold}{\encodingdefault}{\sfdefault}{bx}{n}

\usepackage{amsmath,amssymb}
\usepackage{graphicx}
\usepackage{booktabs}
\usepackage{microtype}
\usepackage{float}
\usepackage{placeins}

\usepackage[normalem]{ulem}
\usepackage{url}
\usepackage{hyperref}
\hypersetup{
  pdftitle={Direct Hidden-State Alignment: Mapping and Controlling Preference Expression in LLMs},
  pdfauthor={Fansheng Zhang, Shengran Guo, Zexiao Wang, Liang Yuan, Jiyuan Chen, Ruikun Luo}
}

\title{Direct Hidden-State Alignment:
Mapping and Controlling Preference Expression in LLMs}

\author{\parbox[t]{0.97\textwidth}{\centering
Fansheng Zhang\textsuperscript{1,*}\quad
Shengran Guo\textsuperscript{2,\dag}\quad
Zexiao Wang\textsuperscript{3,\dag}\\[0.3em]
Liang Yuan\textsuperscript{4,*}\quad
Jiyuan Chen\textsuperscript{1}\quad
Ruikun Luo\textsuperscript{5}\\[0.8em]
\normalfont\small
\textsuperscript{1}Chengdu University\quad
\textsuperscript{2}North Carolina State University\\
\textsuperscript{3}Fudan University\quad
\textsuperscript{4}Australian Catholic University\\
\textsuperscript{5}University of Macau\\[0.5em]
\textsuperscript{\dag}Shengran Guo and Zexiao Wang contributed equally and share second authorship.\\[0.25em]
\textsuperscript{*}Corresponding authors: Liang Yuan and Fansheng Zhang.\\[0.25em]
\href{mailto:liang.yuan@acu.edu.au}{\texttt{liang.yuan@acu.edu.au}}\quad
\href{mailto:above1firmament@gmail.com}{\texttt{above1firmament@gmail.com}}
}}

\iclrfinalcopy

\begin{document}

\maketitle
\fancyhead{} 
\renewcommand{\headrulewidth}{0pt}

\begin{abstract}
In many settings, post-training need not create the target behavior from
scratch: the base model can already produce it, but not reliably. This
shifts part of preference alignment from capability acquisition to
behavioral expression. This raises a central question: if target-supporting
computation is already present, what prevents the preferred behavior from
reliably dominating generation? At each generation step, many Transformer
components write to the same residual stream, yet their effects are combined
into a single next-token distribution. A target-supporting computation can
therefore be present yet be outweighed by other computations. We hypothesize
that reliable expression depends on this internal competition during
inference. We introduce Residual Competition Maps (RCMs), which
map a specified preference onto native residual computation by measuring
signed causal effects relative to that preference. Across preference domains,
RCMs reveal residual competition whose prevalence varies by task, distinct
component roles relative to the target preference, and cases where a single
native-component intervention reverses the preference outcome. DPO
substantially reorganizes these effects and can weaken opposing effects,
which may nevertheless persist. Beyond analysis, RCM provides causal guidance
for where to intervene. To directly control preference formation during
inference, we propose Direct Hidden-State Alignment (DHSA), which
treats inference-time hidden states rather than base-model weights as the
direct adaptation space. RCM-guided Causal Activation State
Transition (CAST) instantiates this principle through local state
interventions at a small number of preference-relevant interfaces while
freezing the base model. With only 256--16,384 controller parameters, CAST
reaches DPO-competitive operating points across three preference domains,
can complement DPO-trained models, and can be enabled or removed at inference
time.
\end{abstract}
\section{Introduction}

\begin{figure}[t]
    \centering
    \includegraphics[width=\linewidth]{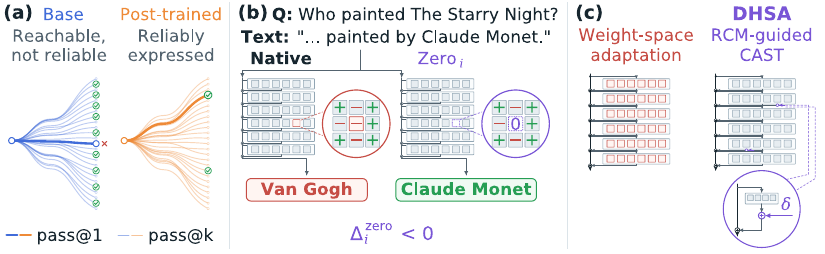}
    \caption{
    \textbf{Overview of RCM-guided direct hidden-state alignment.}
    (a) A schematic sampling contrast illustrates the gap between
    behavioral reachability and reliable expression: bold paths denote
    single draws, while faint trajectories denote candidates exposed
    under a larger sampling budget; trajectory counts are illustrative
    rather than empirical \texttt{pass@k} measurements.
    (b) RCM maps an explicit behavioral preference onto signed causal
    effects of native component computation. Opposing signed effects
    on the same input form the residual competition studied here.
    The schematic example shows a generated preference outcome changing
    after a singleton native-component intervention.
    (c) Model-wide parameter adaptation updates weights across the model.
    In contrast, DHSA directly adapts inference-time hidden states.
    RCM-guided CAST implements this principle at a small number of
    residual interfaces while keeping the base weights frozen.
    }
    \label{fig:overview}
\end{figure}

Preference alignment aims to make desired behavior
reliable~\citep{ouyang2022,rafailov2023}. Yet being able to generate
a target response is different from producing it reliably. LIMA shows
that limited instruction tuning can elicit high-quality outputs,
suggesting that alignment can draw on knowledge acquired during
pretraining~\citep{zhou2023lima}. Recent work on reinforcement learning
with verifiable rewards (RLVR) reports substantial gains in
single-sample success (\texttt{pass@1}) after post-training, while base
models can achieve higher \texttt{pass@k} at larger
$k$~\citep{yue2025}. Here, \texttt{pass@k} measures whether at least
one of $k$ attempts succeeds. These observations distinguish
\emph{behavioral reachability} from \emph{reliable expression}, as
illustrated in Figure~\ref{fig:overview}(a).
\textbf{When a target behavior is already reachable, why can it still
fail to be expressed reliably, and can this failure be corrected
directly?}

To understand this gap, we model generation at the level of the
model's native residual computation. Attention heads, feed-forward modules, and other
components write additive updates to a shared residual
stream~\citep{vaswani2017,elhage2021}. At each generation step, these
contributions combine into a single next-token distribution.
Target-supporting computation can therefore already be present yet fail
to dominate the prediction when other native computations outweigh its
effect. This motivates our hypothesis of \emph{internal competition
during inference}: target-supporting and target-competing native effects
can coexist as a prediction is formed, and their relative influence can
affect which behavior is expressed.

We use \emph{residual competition} for this measurable coexistence of
opposing native effects. For the same input and preference, some
components can causally raise the target preference while others lower
it, as illustrated in Figure~\ref{fig:overview}(b). These effects are defined relative to the specified preference;
they do not assume that component effects form an additive
decomposition of the model output. Under this view, the problem becomes three
sequential questions.
\textbf{First, how can an external behavioral preference be mapped onto
native model computation so that component effects become measurable
relative to it?}
\textbf{Second, what internal structure of preference expression does
this mapping reveal?}
\textbf{Third, can this internal structure directly guide control of
which behavior is expressed?}

The first question requires more than identifying generally important
components. A component can strongly affect generation without
revealing whether it supports or opposes the preference being aligned.
Causal tracing and activation patching have shown how interventions on
internal states can connect model components to their
outputs~\citep{meng2022rome,zhangnanda2024}. We therefore introduce
\emph{Residual Competition Maps} (RCMs), a component-level causal
framework that maps an explicit behavioral preference onto native model
computation. For each input, RCM uses single-component causal contrasts
to measure how each component shifts the target behavior's advantage
over alternatives. The resulting signed effect captures both the
direction and strength of its preference-relative contribution.
Figure~\ref{fig:overview}(b) illustrates this mapping. Because these measurements are causal, RCM can be used not
only to analyze preference expression, but also to guide where to
intervene.

The resulting measurements reveal residual competition across
heterogeneous preference settings, with its prevalence varying
across settings. In representative cases, a single native-component
intervention reverses the generated preference outcome. RCMs further
show that Direct Preference Optimization (DPO) substantially
reorganizes these native effects and can weaken opposing effects,
without guaranteeing their removal. This extends prior mechanistic
evidence that preference post-training reshapes internal
computation~\citep{lee2024}: improved preference behavior need not
eliminate opposing native effects.

Once this structure is measurable, it also provides a direct route to
control. Prior work shows that inference-time state interventions can
steer model behavior without updating base
weights~\citep{rimsky2024,wu2024reft}. To directly control how a
preference is formed during inference, we act on the hidden states that
carry this computation while keeping the base-model weights fixed. We
call this broader principle \emph{Direct Hidden-State Alignment}
(DHSA).

We instantiate DHSA with RCM-guided \emph{Causal Activation State
Transition} (CAST). RCM identifies preference-relevant residual
interfaces, and CAST learns local state changes at a small number of
these sites (Figure~\ref{fig:overview}(c)). Across settings, task-specific scores and controller configurations
instantiate the same RCM-guided state-control principle. This
gives a low-footprint form of alignment without reparameterizing the
base model. Across heterogeneous preference settings and model
families, CAST reaches DPO-competitive operating points with only
256--16,384 controller parameters. The controllers remain separate
from the base weights, can be enabled or removed at inference time,
and can also improve DPO-trained models.

Our main contributions are:
\begin{enumerate}
    \item \textbf{Component-level preference mapping.}
    We introduce RCMs, which map an external behavioral preference
    onto signed causal effects of native model computation through
    single-component contrasts. RCM measures whether and how strongly
    each component shifts the target behavior's advantage over
    alternatives, providing both an analysis tool and causal guidance
    for intervention.

    \item \textbf{Residual competition and preference-relative
    component roles.}
    Across heterogeneous preference settings, RCMs reveal coexisting
    target-supporting and target-competing native effects and
    input-dependent component roles. In representative cases, a
    single native-component intervention reverses the generated
    preference outcome. DPO substantially reorganizes these effects
    and can weaken opposition without guaranteeing its removal.

    \item \textbf{Direct control of preference formation.}
    We introduce DHSA to directly adapt the hidden states through
    which preferences are expressed during inference, while keeping
    base-model weights frozen. RCM-guided CAST provides a simple
    low-footprint implementation at a small number of
    preference-relevant interfaces. With 256--16,384 controller
    parameters, CAST reaches DPO-competitive operating points across
    the studied settings, can further improve DPO-trained models, and
    remains detachable at inference time.
\end{enumerate}
\section{Related Work}

\subsection{Preference Alignment and Behavioral Expression}

Preference alignment adapts language models using feedback about desired
behavior. RLHF optimizes a policy with learned preference
rewards~\citep{ouyang2022}, while DPO learns directly from preference
pairs~\citep{rafailov2023}. Beyond optimization algorithms,
task-complexity analyses examine the additional information needed to
access pretrained capabilities~\citep{vergarabrowne2026}. RLVR studies
investigate sampling coverage~\citep{yue2025}, while reasoning-aware
evaluations assess intermediate-step correctness and report expanded
reasoning capabilities~\citep{wen2026}. These perspectives examine
different aspects of post-training, from behavioral accessibility to
reasoning quality. Our work focuses on behavioral expression: how a
specified preference appears in native model computation, how
post-training reorganizes that computation, and whether the same
internal structure can guide direct control during inference.

\subsection{Mechanistic Interpretability and Behavioral Attribution}

Causal tracing, activation patching, and circuit analysis connect model
behavior to internal components~\citep{meng2022rome,wang2022ioi}.
Benchmarks and methodological studies examine the causal efficacy of
interpretability methods and the sensitivity of localization to
evaluation choices~\citep{arora2024causalgym,zhangnanda2024}.
Studies of refusal identify a low-dimensional control
direction~\citep{arditi2024}, while mechanistic analysis of DPO finds
that toxicity reduction can bypass pretrained
capabilities~\citep{lee2024}. GCM uses contrastive-response likelihoods
to select causal mediators for sparse
steering~\citep{sankaranarayanan2026}; Preference Heads combines causal
head masking with contrastive decoding for
personalization~\citep{zhang2026preferenceheads}. Subspace-patching
studies distinguish successful behavioral manipulation from faithful
attribution~\citep{makelov2024subspace}. These studies connect causal
attribution with behavioral control. RCM complements this line by
mapping a specified preference onto signed causal effects of native
computation, supporting both mechanistic analysis and intervention
guidance.

\subsection{Representation-Based Control and Alignment}

Activation engineering and representation engineering develop internal
representations as behavioral control
interfaces~\citep{turner2023actadd,zou2023repe}. ITI and CAA demonstrate
inference-time steering through localized or contrastively constructed
activation interventions~\citep{li2023iti,rimsky2024}. Conditional
Activation Steering, also abbreviated CAST, gates steering using
input-dependent activations~\citep{lee2025conditional}; our CAST learns
state changes at RCM-selected interfaces. ReFT learns interventions on
frozen-model representations, with LoReFT using low-rank
parameterization; LoFiT learns task-specific attention-head selection
followed by additive offsets at selected
heads~\citep{wu2024reft,yin2024lofit}. Preference-oriented approaches
further extend representation control to alignment: RAHF manipulates
preference-related representations~\citep{liu2024rahf}, RE-Control
formulates alignment as state-space control~\citep{kong2024recontrol},
and BiPO learns steering vectors from pairwise
preferences~\citep{cao2024bipo}. PaLRS extracts training-free residual
steering vectors from preference pairs and reports performance
competitive with or better than DPO and
SimPO~\citep{meng2024simpo} on mathematical reasoning and code
generation~\citep{lacava2025}. These studies establish that
representation-level control can support preference alignment and can
reach strong alignment baselines.

We use Direct Hidden-State Alignment (DHSA) to frame preference
adaptation at the level of inference-time hidden states of a frozen
base model. RCM provides preference-guided localization by identifying
preference-relevant native effects, and CAST applies local state changes
at a small number of those interfaces. We evaluate this RCM-guided
realization of DHSA across heterogeneous preference settings.
\section{Method}
\label{sec:method}

\subsection{How Preference Appears in Token Predictions Formed by Component Writes}
\label{sec:advantage}

In the residual language models studied here, the residual stream
carries the shared state used to predict each next token.
Attention and MLP components read from this stream and write their
outputs back to it. Later components read the updated state.
The state at the current prediction position is ultimately mapped
to the next-token distribution
\citep{vaswani2017,elhage2021}.

Let $p$ be a token position and $r_p\in\mathbb R^d$ its residual
state before a component write.
For component $i$, let $h_{p,i}\in\mathbb R^{d_i}$ be its native
(unintervened) output and $P_i\in\mathbb R^{d\times d_i}$ its fixed
write-back map.
The component changes the shared state by
\begin{equation}
    z_{p,i}=P_i h_{p,i},
    \qquad
    r_p\leftarrow r_p+z_{p,i}.
    \label{eq:residual_write}
\end{equation}
For an attention head, $P_i$ is its slice of the attention output
projection. Through the residual stream, native component writes
affect later computation and the tokens the model produces.

Each experiment specifies a preference $e$: a target behavior
relative to a non-target behavior.
We quantify its expression using established, widely used forms
of task scoring.
Write the resulting target advantage for model computation $M$
on input $x$ as $C_e(M;x)$, with larger values favoring the target.
Appendix~\ref{app:task_scores} specifies its implementation for
each task in our experiments.

The score connects the external preference to the model's output.
Component writes connect native computation to that output.
RCM maps the preference onto the causal effects of these writes.

\subsection{How RCM Maps Preference to Native Component Effects}
\label{sec:rcm}

RCM uses a single component as its unit of comparison and evaluates
the resulting change with the specified preference score.

For component $i$, let $a$ and $b$ specify two state conditions at
the intervention positions.
Let $M_i^a$ and $M_i^b$ be the resulting computations, including
all downstream changes.
For the same input $x$, RCM measures
\begin{equation}
    \Delta_{i,e}^{a\rightarrow b}(x)
    =
    C_e(M_i^b;x)-C_e(M_i^a;x).
    \label{eq:rcm_effect}
\end{equation}
Repeating this single-component comparison maps the externally
specified preference onto signed causal effects of native
component states.

We use two instances of this general component-state contrast
in our experiments. RCM-Zero compares a zeroed write with its
native write. A positive effect means the native component raises
the target advantage; a negative effect means it lowers it.
Figure~\ref{fig:overview}(b) illustrates this
native-versus-intervened contrast.
RCM-Patch compares the receiving state with a specified source
state. It measures what that state replacement does to the same
advantage. Appendix~\ref{app:rcm_patch} specifies the source
conditions.

The signs give components a preference-relative meaning.
For one input, \emph{residual competition} occurs when scanned
components include both target-raising and target-lowering
RCM-Zero effects above the stated thresholds.
Across inputs, we name components with positive mean effects
\emph{target-supporting}, those with negative mean effects
\emph{target-competing}, and those with substantial effects of
both signs \emph{bidirectional regulatory}. These descriptive roles
can overlap.
Appendix~\ref{app:roles} gives the role criteria.

\subsection{How Native Effects Guide Direct Hidden-State Control}
\label{sec:cast}

An RCM effect identifies a component state whose change causally
moves the measured target advantage.
Its location and sign describe how the model's native computation
supports or opposes the specified preference.
Residual competition further shows where opposing effects coexist.
The resulting map therefore provides a causal basis for direct
control of preference expression.

Following the DHSA principle introduced in Section~1, we instantiate
this control with RCM-guided \emph{Causal Activation State Transition}
(CAST).
CAST learns local changes to component states at locations selected
using RCM.
A selection may use target-supporting locations,
target-competing locations, or both.
In the experiments, we scan coarse components before attention
heads and use effect estimates with confidence intervals to rank
candidates. Appendix~\ref{app:selection} gives the tested
selection rules and budgets.

Let $\mathcal G$ be the selected locations.
For $i\in\mathcal G$, let
$h^{\mathrm{ctrl}}_{p,i}\in\mathbb R^{d_i}$ be the component state
in the current controlled forward pass.
The controller
$g_i(\,\cdot\,;\theta_i):\mathbb R^{d_i}\to\mathbb R^{d_i}$
returns a state change, with trainable parameters $\theta_i$.
The mask $m_p\in\{0,1\}$ selects token positions, and
$\alpha\geq0$ sets the intervention strength.
CAST applies the resulting state change before the residual write:

\begin{equation}
    \widetilde h_{p,i}
    =
    h^{\mathrm{ctrl}}_{p,i}
    +\alpha m_p\,g_i(h^{\mathrm{ctrl}}_{p,i};\theta_i),
    \qquad
    \widetilde z_{p,i}=P_i\widetilde h_{p,i}.
    \label{eq:cast_control}
\end{equation}
Here $\widetilde h_{p,i}$ is the transformed component state and
$\widetilde z_{p,i}$ is its residual write.
The local state change in Figure~\ref{fig:overview}(c) corresponds to
$\alpha m_p\,g_i(h^{\mathrm{ctrl}}_{p,i};\theta_i)$.
In our experiments, the controller is either a constant vector
$g_i(h;\theta_i)=v_i$, with $v_i\in\mathbb R^{d_i}$,
or a low-rank map
$g_i(h;\theta_i)=B_iA_i h/\sqrt{k_i}$.
The latter has rank $k_i$,
$A_i\in\mathbb R^{k_i\times d_i}$, and
$B_i\in\mathbb R^{d_i\times k_i}$.
The model then continues its forward computation from the
transformed states.
When several locations are selected, we evaluate their joint
effect on the model's output.

We denote the constant-vector and low-rank variants by CAST-SV and
CAST-LR. The \texttt{prefill} configuration acts at registered prompt
positions; \texttt{all} also acts during decoding.
Appendix~\ref{app:timing} specifies their training and inference
masks.

\subsection{How the Control Change Is Learned}
\label{sec:learning}

Let $\theta$ collect the trainable controllers at the selected
locations, and let $\omega$ denote the fixed base-model weights.
CAST specifies how $\theta$ changes inference states; the learning
rule is separate from the state transformation. In this paper, we
evaluate a reference-relative pairwise objective of the DPO form.
The same learning form is used across the evaluated tasks; only the
task-specific response score and controller configuration differ.

Our experiments use preference pairs $(x,y^+,y^-)$.
Training increases the preferred response's score relative to
the non-preferred response and to the same model with control
disabled.
The reference-relative pairwise loss, response scores, and
regularization are specified in
Appendices~\ref{app:training} and~\ref{app:task_scores}.
Appendix~\ref{app:advantage_equivalence} derives the relation of
this experimental loss to DPO's pairwise form
\citep{rafailov2023}.

\subsection{Direct Hidden-State Alignment}
\label{sec:dhsa}

DHSA describes the broader adaptation principle behind CAST:
preference alignment is performed by transforming hidden states
during the model's inference computation while keeping the
base-model weights fixed.

Let $H_t$ denote a collection of hidden states reached while the
model predicts token $t$, and let $\phi$ parameterize a state
transformation. DHSA acts as
\begin{equation}
    \widetilde H_t
    =
    T_\phi(H_t;x,y_{<t}).
    \label{eq:dhsa_transform}
\end{equation}
The model continues its computation from $\widetilde H_t$ to
produce the next-token distribution.
The transformation can act at different states, positions, and
granularities. CAST is the sparse RCM-guided instance studied here:
it restricts the transformation to a small number of
preference-relevant interfaces while leaving the base weights unchanged.
Appendix~\ref{app:local_controllability} analyzes when a local
hidden-state change improves a preference objective.
\section{Experiments}
\label{sec:experiments}

We first use RCM to characterize preference-relative native effects,
then test whether the resulting maps guide localized control, and
finally evaluate RCM-guided hidden-state alignment across three
preference settings.
\footnote{\url{https://github.com/abovefiramament/DHSA}}

\subsection{Experimental Setup}
\label{sec:exp_setup}

\paragraph{Tasks and models.}
The tasks cover distinct demands on preference control.
ConFiQA~\citep{bi2025contextdpo} simulates retrieval-augmented factual
answering under conflict between supplied evidence and parametric
knowledge; PC measures target-answer occurrence, and EM requires the
entire normalized output to match a context-supported answer alias.
IMDb~\citep{maas2011imdb} tests the precision of sentiment alignment:
reward measures movement toward the desired sentiment, alongside
measures of distribution shift. TL;DR tests an open-generation
strategy in which the model must select and compress content into a
preferred summary~\citep{volske2017tldr,stiennon2020summarize}.
Appendix~\ref{app:evaluation} gives the evaluation definitions.
Across tasks, preference scores and controller configurations
instantiate the same RCM-guided state-alignment principle.

Site experiments use Llama-3-8B and Qwen2.5-14B on ConFiQA,
GPT-2-large and Qwen2.5-14B on IMDb, and GPT-J-6B and Qwen2.5-14B
on TL;DR. Performance experiments use Llama-3-8B and Qwen2-7B
on ConFiQA, GPT-2-large on IMDb, and GPT-J-6B and LION-Llama-3-8B
on TL;DR. Starting checkpoints are instruction-tuned or SFT models.

Site test sizes are 2,048 per ConFiQA subset for Llama-3-8B;
500/478/477 for Qwen2.5-14B QA/MR/MC; 2,048 per IMDb model;
and 2,048/320 for GPT-J/Qwen TL;DR. Performance uses 2,048 mixed
ConFiQA inputs per model, 2,048 IMDb prompts, and 512 TL;DR inputs
per model. TL;DR preference is judged against human references and
averaged over both presentation orders.

\paragraph{Comparisons and budgets.}
Released DPO checkpoints provide representative preference-alignment
references. DPO serves as an evaluation-matched reference; training
data and compute are not matched across methods. BiPO and LoReFT use
their released implementations and native objectives. Candidate-training
opportunities, active parameter counts, and available runtime
measurements are reported separately. Site compares RCM-Zero,
RCM-Patch, ITI, and three fixed random selectors under the registered
selection and deployment protocols. Validation data select intervention
strength where a single operating point is reported; IMDb retains the
measured reward--KL curves. Performance reports both \texttt{all} and
\texttt{prefill} control. Full settings and costs appear in
Appendices~\ref{app:cast_protocol}--\ref{app:implementation}.

\subsection{Residual Competition and Component Reorganization}
\label{sec:residual_competition_results}

We measure RCM-Zero effects for 32 Llama-3-8B attention-layer
components on 300 ConFiQA inputs. Each effect is the
native-minus-ablated change in paired target--non-target
token-logit margin. An input has bilateral effects at threshold
$\tau$ when one component exceeds $\tau$ and another falls below
$-\tau$; $\tau$ is an absolute change in margin units.

\begin{figure}[!htbp]
    \centering
    \includegraphics[width=\linewidth]
        {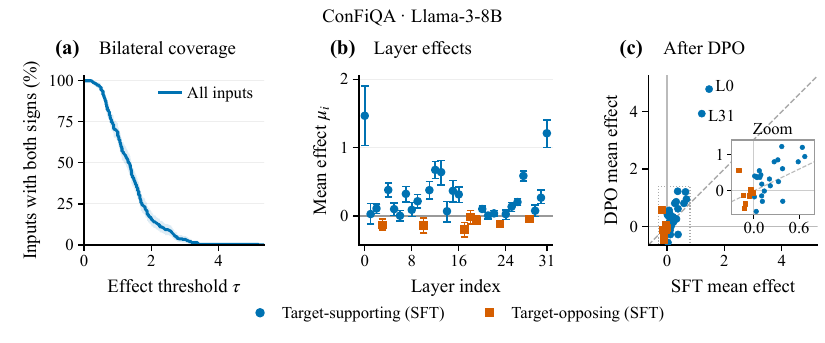}
    \caption{
    \textbf{Residual competition in ConFiQA.}
    RCM-Zero effects across 32 Llama-3-8B attention layers on
    300 inputs. (a) Bilateral coverage against the absolute
    token-logit-margin threshold. (b) Mean effects with 95\%
    confidence intervals. (c) Paired SFT--DPO mean effects.
    Colors and markers indicate the SFT mean-effect sign.
    }
    \label{fig:rcm_results}
\end{figure}

At $\tau=0.5$, bilateral effects occur in 283/300 inputs
(Figure~\ref{fig:rcm_results}(a)).
Target output need not imply uniformly target-supporting component
effects: Appendix~\ref{app:target_outputs} gives native target answers
with opposing effects, and Appendix~\ref{app:singleton_cases} gives
actual decoded changes under single-head ablation.
Figure~\ref{fig:rcm_results}(b) summarizes signed means;
Appendix~\ref{app:roles} distinguishes them from substantial per-input
effects that cancel in the mean.

DPO changes layer effects by 1.00 margin units on average in
absolute value (95\% CI: 0.97--1.04), and 8/32 layers change
mean-effect sign (Figure~\ref{fig:rcm_results}(c)).
Yet 11 layers retain negative mean effects, 8 with pointwise
95\% intervals below zero. DPO therefore substantially reorganizes
the measured effects without guaranteeing the removal of
target-opposing computation.

Appendix~\ref{app:thresholds} extends the analysis to IMDb and TL;DR.
Bilateral effects occur in 28/300 IMDb inputs at $\tau=0.01$ and
289/300 TL;DR inputs at $\tau=0.02$, in each task's own score units.
The prevalence of competition therefore differs across tasks.

\FloatBarrier

\subsection{Does RCM Guide Effective Localized Control?}
\label{sec:site_results}

Site experiments test whether preference-relative RCM measurements
provide useful locations for subsequent hidden-state control.
The unit of comparison is the registered localization-and-control
configuration, including each selector's bank organization.
Vector configurations use RCM's separate signed banks, whereas ITI
and Random use their registered joint banks. For TL;DR, all selectors
receive the same single-head training and audit procedure before
eight-head composition (Appendix~\ref{app:baselines}).

Table~\ref{tab:site_main} shows that RCM-Zero exceeds every random
selector across both ConFiQA and TL;DR model families. ITI leads on
ConFiQA Llama-3-8B, whereas RCM-Zero leads on ConFiQA Qwen2.5-14B
and both TL;DR models. The TL;DR advantage is particularly clear;
RCM-Patch is less consistent across models. RCM measurements therefore provide actionable localization, with
the strongest selector depending on the model and setting.

\begin{table}[!htbp]
    \centering
    \small
    \setlength{\tabcolsep}{5pt}
    \begin{tabular}{lrrrr}
        \toprule
        & \multicolumn{2}{c}{ConFiQA PC $\uparrow$}
        & \multicolumn{2}{c}{TL;DR Win $\uparrow$} \\
        \cmidrule(lr){2-3}
        \cmidrule(lr){4-5}
        Localization
        & Llama-3-8B & Qwen2.5-14B
        & GPT-J-6B & Qwen2.5-14B \\
        \midrule
        Random 1
        & 63.74 & 61.35 & 45.17 & 48.59 \\
        Random 2
        & 62.48 & 68.30 & 50.73 & 48.59 \\
        Random 3
        & 66.08 & 74.89 & 46.83 & 48.28 \\
        ITI
        & \textbf{85.64} & 75.15 & 46.83 & 50.16 \\
        RCM-Patch
        & 80.40 & 72.75 & 53.88 & 48.91 \\
        RCM-Zero
        & 78.61 & \textbf{83.53}
        & \textbf{54.64} & \textbf{56.09} \\
        \bottomrule
    \end{tabular}
    \caption{
    \textbf{Site results} (\%).
    ConFiQA reports macro-average PC across QA, MR, and MC;
    TL;DR reports order-balanced preference against human
    references. All three fixed random selectors are shown.
    }
    \label{tab:site_main}
\end{table}
\FloatBarrier

\begin{figure}[!htbp]
    \centering
    \includegraphics[width=\linewidth]
        {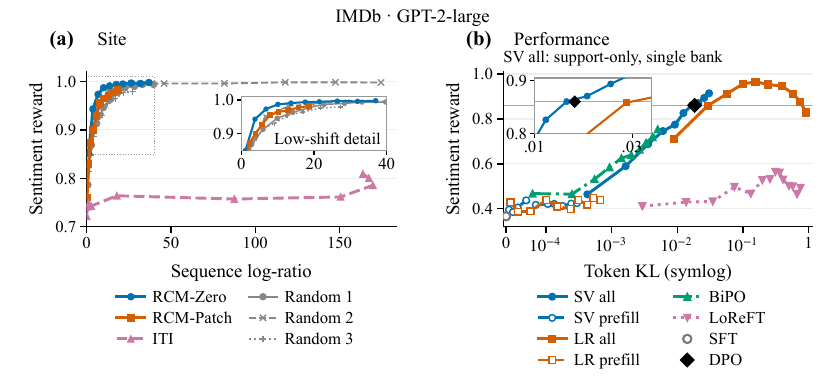}
    \caption{
    \textbf{IMDb localization and control efficiency.}
    GPT-2-large, 2,048 test prompts per protocol.
    (a) Site reward--sequence-log-ratio curves.
    (b) Performance reward--token-KL curves.
    SV/LR denote CAST's vector/low-rank controllers; LoReFT is the
    external baseline. SV-all uses the supporting-only bank;
    the black diamond denotes DPO.
    }
    \label{fig:imdb_results}
\end{figure}

\paragraph{IMDb.}
Figure~\ref{fig:imdb_results}(a) tests whether RCM-guided
localization improves sentiment reward without excessive
sequence-level shift. The same signed RCM candidate rule identifies
weak competing-side effects in this setting. The main CAST-SV-all
efficiency comparison therefore deploys only the four
target-supporting heads; the candidate-selection rule itself is
unchanged. Appendix~\ref{app:imdb_site} reports the additional
Qwen2.5-14B Site result.

\FloatBarrier

\subsection{Can Direct Hidden-State Alignment Reach DPO-Competitive Operating Points?}
\label{sec:performance_results}

We next ask whether DHSA can reach the behavioral regime of
preference post-training across distinct preference settings. Table~\ref{tab:main_performance} compares RCM-guided CAST
with DPO, BiPO, and LoReFT. CAST uses the task-appropriate controller
form specified in Section~\ref{sec:cast}: constant vectors on
ConFiQA and low-rank interventions on TL;DR. We also test controllers
transferred from SFT to DPO (\textbf{T}) and controllers trained
directly on DPO (\textbf{NT}).

\begin{table}[!htbp]
    \centering
    \small
    \setlength{\tabcolsep}{3.4pt}
    \begin{tabular}{lrrrrrr}
        \toprule
        & \multicolumn{4}{c}{ConFiQA}
        & \multicolumn{2}{c}{TL;DR} \\
        \cmidrule(lr){2-5}
        \cmidrule(lr){6-7}
        & \multicolumn{2}{c}{Llama-3-8B}
        & \multicolumn{2}{c}{Qwen2-7B}
        & GPT-J & LION \\
        Method
        & PC $\uparrow$ & EM $\uparrow$
        & PC $\uparrow$ & EM $\uparrow$
        & Win $\uparrow$ & Win $\uparrow$ \\
        \midrule
        SFT
        & 59.42 & 3.71
        & 39.26 & 0.15
        & 44.43 & 71.39 \\
        DPO
        & 88.92 & 43.99
        & 79.30 & 3.86
        & 62.11 & 84.67 \\
        BiPO
        & 81.98 & 70.80
        & 84.91 & 75.15
        & 51.86 & 78.03 \\
        LoReFT
        & 82.96 & 52.05
        & 89.94 & 77.25
        & 47.07 & 67.48 \\
        CAST (all)
        & 81.93 & 0.00
        & 82.96 & 16.31
        & 61.72 & 79.10 \\
        CAST (prefill)
        & 82.52 & 64.65
        & 66.75 & 24.95
        & 52.83 & 82.81 \\
        \midrule
        DPO+CAST-T (all)
        & 88.92 & 47.95
        & 89.31 & 39.84
        & 68.55 & 85.55 \\
        DPO+CAST-T (prefill)
        & 90.38 & 39.31
        & 83.50 & 39.50
        & 65.04 & 87.01 \\
        DPO+CAST-NT (all)
        & 88.33 & 57.23
        & 83.89 & 8.59
        & 63.96 & 82.62 \\
        DPO+CAST-NT (prefill)
        & 90.53 & 55.52
        & 82.28 & 27.49
        & 61.82 & 84.67 \\
        \bottomrule
    \end{tabular}
    \caption{
    \textbf{Preference-alignment performance} (\%).
    CAST uses constant vectors on ConFiQA and low-rank controllers on TL;DR.
    TL;DR Win is against human references, not directly against DPO.
    T transfers an SFT-trained controller to DPO and recalibrates
    $\alpha$ without retraining; NT trains on the DPO checkpoint.
    }
    \label{tab:main_performance}
\end{table}

\paragraph{ConFiQA: factual inference under conflict.}
CAST enters the DPO-competitive range on context-supported
answering, but PC and EM expose different outcomes. On Llama-3-8B,
prefill control improves exact answers beyond DPO while remaining
lower on PC; all-position control raises PC but fails on EM.
On Qwen2-7B, CAST-all exceeds DPO on both metrics, while BiPO
and LoReFT remain stronger. The contrast also shows that
intervention timing changes how the same preference is expressed.

\paragraph{IMDb: precision of sentiment alignment.}
Figure~\ref{fig:imdb_results}(b) compares sentiment reward at
the accompanying token KL. The target-supporting CAST bank reaches
the DPO reward with lower measured KL (approximately 0.0166
versus 0.0183 by interpolation). BiPO and LoReFT do not reach
that reward within their evaluated grids. In this setting,
localized hidden-state adaptation reaches the DPO reward at a
smaller measured token KL.

\paragraph{TL;DR: open-generation strategy.}
CAST approaches DPO on both model families and exceeds the
corresponding BiPO and LoReFT results. This extends hidden-state
alignment to open-ended generation, where preference depends on
content selection and compression across a complete response.

\paragraph{External baselines.}
Across the three settings, CAST reaches DPO-competitive operating
points with task- and metric-dependent tradeoffs; BiPO and LoReFT
remain particularly strong on ConFiQA.

\paragraph{Compatibility with DPO.}
Transferred CAST yields higher observed TL;DR win rates on both
models and improves several ConFiQA metrics, while controllers
trained directly on DPO yield mixed gains. IMDb shows the same
compatibility with a reward--KL tradeoff
(Appendix~\ref{app:imdb_dpo}). Thus DHSA can complement parameter-level preference optimization.
Paired uncertainty is reported in Appendix~\ref{app:performance_ci}.
\section{Discussion and Conclusion}
\label{sec:discussion}

RCM maps specified preferences onto signed causal effects of native
residual computation, revealing residual competition and
preference-relative component roles. Opposing effects can persist
under preferred outputs and after substantial reorganization by
preference post-training. RCM-guided CAST shows that the resulting
causal structure can guide localized control. With frozen base
weights, CAST realizes DHSA as a direct adaptation route through
inference-time hidden states that can also complement parameter-level
preference optimization. This connects a measurable account of preference expression to a concrete
intervention space, linking mechanistic analysis with direct state control.

The main limitations are the evaluated preferences, model families,
intervention settings, and our empirical focus on reference-relative
pairwise training. Future work can study adaptive state control and
other learning signals. Overall, when a target behavior is already
reachable, its expression can be analyzed and adjusted directly
through the inference-time computation that shapes generation.
\subsection*{AI Use Statement}

We used generative AI tools for manuscript proofreading,
literature retrieval and discovery, and clarification of conceptual
and methodological arguments.
The authors are responsible for verifying the cited sources,
mathematical arguments, and reported results, and
take responsibility for the final content and AI-assisted materials.

\subsection*{Ethics Statement}

This work studies how internal model computation supports or opposes
specified preferences and how local state interventions can alter
model behavior. Such control has potential dual uses: it may support
desirable behavior, but could also be used to weaken safety constraints
or steer outputs toward harmful objectives.
The preference metrics studied here measure task-specific behavior
and do not constitute comprehensive safety assessments.
Deployment therefore requires evaluation of unintended behavioral
changes, appropriate safeguards, and compliance with the licenses
and usage conditions of the underlying models and datasets.

\subsection*{Reproducibility Statement}

Section~\ref{sec:method} specifies the RCM measurements and CAST
interventions. Appendices~\ref{app:task_scores},
\ref{app:selection}, and~\ref{app:training} describe the task-specific
scores, component-selection procedures, and pairwise training
objective used in our experiments.
The assumptions and proofs underlying the theoretical results are
provided in Appendices~\ref{app:advantage_equivalence},
\ref{app:local_controllability}, and~\ref{app:rcm_proxy}.
Appendix~\ref{app:implementation} reports the experimental settings
and evaluation definitions. The artifact repository is
linked in Section~\ref{sec:experiments}.

\bibliographystyle{iclr2027_conference}
\bibliography{references}

\appendix

\section{Task Scores and CAST Protocols}
\label{app:cast_protocol}

\subsection{Task-Specific Preference Scores}
\label{app:task_scores}

The main text uses $C_e(M;x)$ as a generic preference score and
$s_{e,\theta}(x,y)$ as the differentiable response score used for controller
training. These quantities share the same preference direction but are not
required to be numerically identical.

Let $\ell_{\theta,t}\in\mathbb{R}^{|\mathcal V|}$ denote the next-token logits
of the controlled model at response position $t$, where $\mathcal V$ is the
vocabulary. Let $\mathcal T(y)$ denote the token positions of response $y$.
We use two response scores.

\paragraph{Mean token-logit margin.}
For response $y$,
\begin{equation}
s_{\theta}^{\mathrm{margin}}(x,y)
=
\frac{1}{|\mathcal T(y)|}
\sum_{t\in\mathcal T(y)}
\left[
\ell_{\theta,t}(y_t)
-
\max_{v\in\mathcal V,\;v\neq y_t}
\ell_{\theta,t}(v)
\right].
\label{eq:app_margin_score}
\end{equation}
The score measures the average logit advantage of the designated response token
over its strongest token-level alternative.

\paragraph{Mean response-token log-probability.}
For response $y$,
\begin{equation}
s_{\theta}^{\mathrm{logp}}(x,y)
=
\frac{1}{|\mathcal T(y)|}
\sum_{t\in\mathcal T(y)}
\log
\pi_{\theta}(y_t\mid x,y_{<t}).
\label{eq:app_logp_score}
\end{equation}
The response is teacher-forced while this score is evaluated. The normalization
by response length is part of the score definition and differs from an
unnormalized sequence log-probability.

\paragraph{ConFiQA.}
For the Performance scan, a registered context-supported response $y^+$ and
competing response $y^-$ define the RCM measurement score
\begin{equation}
C_e(M;x)
=
s^{\mathrm{margin}}(M;x,y^+)
-
s^{\mathrm{margin}}(M;x,y^-).
\label{eq:app_confiqa_rcm}
\end{equation}
The Site scan uses the same per-response token-logit margin but takes the
maximum score over the admitted answer aliases separately on each side, in
each model state being compared. Performance scores its registered single
chosen/rejected endpoint pair. Thus the preference direction and token score
agree, while the endpoint aggregation need not give identical numerical
effects when multiple aliases are admitted.
CAST training uses the corresponding mean token-logit-margin response score in
Equation~\ref{eq:app_margin_score}. Final evaluation is performed on generated
answers rather than on this training surrogate and reports the task-specific
generation metrics described in the experimental section.

\paragraph{IMDb.}
RCM measurement is performed on generated continuations. Let
$\widehat y_M(x)$ denote a continuation generated by model $M$ under the fixed
decoding protocol. The frozen sentiment classifier assigns
$R_e(y)=P(\mathrm{POSITIVE}\mid y)-P(\mathrm{NEGATIVE}\mid y)\in[-1,1]$;
the RCM score is
\begin{equation}
C_e(M;x)
=
R_e(\widehat y_M(x)).
\label{eq:app_imdb_rcm}
\end{equation}
When multiple generations are used, Equation~\ref{eq:app_imdb_rcm} is replaced
by their empirical mean. CAST training does not backpropagate through the
sentiment reward; it instead uses
$s^{\mathrm{logp}}_{\theta}(x,y)$ from
Equation~\ref{eq:app_logp_score} on fixed preference pairs. Final evaluation
again uses generated continuations and reports external sentiment reward
together with distribution-shift and generation-health measures.

\paragraph{TL;DR.}
For a preferred summary $y^+$ and rejected summary $y^-$, RCM uses
\begin{equation}
C_e(M;x)
=
s^{\mathrm{logp}}(M;x,y^+)
-
s^{\mathrm{logp}}(M;x,y^-),
\label{eq:app_tldr_rcm}
\end{equation}
and CAST training uses the same form of mean response-token log-probability.
Final evaluation is performed on generated summaries through pairwise
preference evaluation rather than through the training score.

\paragraph{Measurement, optimization, and evaluation.}
The three stages are intentionally distinguished:
\begin{equation}
\text{RCM measurement}
\;\neq\;
\text{CAST training surrogate}
\;\neq\;
\text{final generation metric}
\label{eq:app_score_stages}
\end{equation}
in general. What is shared is the target preference $e$, not necessarily one
numerical objective. Intervention positions and response-scoring positions are
also distinct: an intervention may modify prompt or response states while the
training response score is computed only over its designated response-token
positions.
RCM effects are singleton differences between complete model computations;
they are not assumed to add across components. Native residual writes and CAST
state updates are additive at their local application sites, but downstream
computation can make a joint score change differ from the sum of singleton
RCM effects.

\subsection{RCM Candidate Selection}
\label{app:selection}

RCM selection proceeds hierarchically from coarse components to attention
heads. All selection statistics are computed from per-example singleton RCM
effects.

For candidate component $i$, let
\begin{equation}
\mu_i
=
\frac{1}{N}
\sum_{n=1}^{N}
\Delta_{i,e}^{\mathrm{zero}}(x_n),
\label{eq:app_rcm_mean}
\end{equation}
where $\{x_n\}_{n=1}^{N}$ is the RCM discovery set. Let
$[L_i,U_i]$ denote the corresponding $95\%$ confidence interval computed over
the paired per-example effects.

The high and low sides use the directional confidence scores
\begin{equation}
q_i^{\mathrm{high}}=L_i,
\qquad
q_i^{\mathrm{low}}=-U_i.
\label{eq:app_directional_ci}
\end{equation}
Signed selections admit candidates to these sides by the sign of their own
mean effect, so the bounds favor large effects stable in the corresponding
direction. IMDb Performance instead ranks disjoint high- and low-effect
extrema by the same bounds without requiring the low-end mean to be negative.
An interval crossing zero does not itself exclude a candidate.

\paragraph{Hierarchical scan.}
Selection uses the following sequence:
\begin{enumerate}
    \item scan the registered layer- or component-level candidate set;
    \item form the registered signed roles or disjoint high/low extrema;
    \item rank the high/positive side by $L_i$ and the low/negative side by $-U_i$;
    \item retain the selected parent layers;
    \item scan all attention heads contained in the union of those layers;
    \item assign and rank every head using the head's own effect.
\end{enumerate}
A head therefore does not inherit the sign of its parent layer.
The coarse unit is a pre-output-projection attention layer. Four parent
layers are retained on each signed side or effect extreme, and every head in
their union is scanned before the final head ranking. Scanning acts
on scored decision states. ConFiQA and TL;DR use the teacher-forced prefixes
of their fixed answer or summary endpoints; IMDb uses the prefixes of each
actually generated continuation. In either case, the final non-padding
prompt state predicts the first output token and each subsequent prefix-final
state predicts the next token. Padding and post-EOS states are excluded.
Native and intervened trajectories use this same position rule, even when
their generated tokens differ.

The candidate pool is fixed before actuator training. The registered
ConFiQA and IMDb vector Site comparisons select exactly four positive-effect
and four negative-effect heads ($K=8$). Confidence intervals determine the
within-role ranking; an interval crossing zero does not exclude a candidate
or change the head budget.

IMDb Performance selects four heads at each CI-ranked effect extreme instead;
its low-effect group can contain heads with positive mean effects and is not
necessarily a competing-side group. Its supporting-only deployment activates
the four high-effect heads and leaves the weak low-effect bank inactive.

\paragraph{Low-rank configuration.}
For the more complex TL;DR open-generation task, a larger signed candidate
pool is formed before training. The 16 candidate single-head controllers are
trained and audited individually on 40 non-test posts, against the human
reference in both presentation orders. The final eight-head control set is
assembled from the audit without retraining the selected controllers jointly.
Thus the \emph{candidate pool} is determined before training, whereas the
final audited combination is determined afterward.

\paragraph{Confidence intervals.}
The reported RCM mean-effect intervals use paired percentile bootstrap over
the 300 discovery inputs (500 resamples, seed 42, pointwise 2.5th and 97.5th
percentiles). The interval is for the mean singleton contrast, not for the
probability that an individual input has a nonzero effect; it is not corrected
for multiplicity across scanned components. No test-set metric is used to
alter the RCM candidate pool.

\subsection{Pairwise Training Used in Our Experiments}
\label{app:training}

This section specifies the reference-relative pairwise objective and
training settings used in our experiments.

For a training triple $(x,y^+,y^-)$, let
$s_{e,\theta}(x,y)$ denote the task-specific differentiable response score under
CAST, and let $s_{e,0}(x,y)$ denote the score from the same frozen starting
model with the controller disabled. Define
\begin{equation}
d_{e,\theta}(x)
=
s_{e,\theta}(x,y^+)
-
s_{e,\theta}(x,y^-)
\label{eq:app_pair_advantage}
\end{equation}
and the reference-relative advantage gain
\begin{equation}
A_{e,\theta}(x)
=
d_{e,\theta}(x)
-
\operatorname{stopgrad}\!\left(d_{e,0}(x)\right).
\label{eq:app_advantage_gain}
\end{equation}
Here $\operatorname{stopgrad}(\cdot)$ evaluates its argument numerically but
blocks gradients through it.

For the experiments reported in this paper, we minimize
\begin{equation}
\mathcal L_{\mathrm{CAST}}(\theta)
=
-
\mathbb E_{(x,y^+,y^-)\sim\mathcal D_{\mathrm{train}}}
\log\sigma\!\left(\beta A_{e,\theta}(x)\right)
+
\lambda
\sum_{W\in\theta}
\frac{\|W\|_F^2}{|W|},
\label{eq:app_cast_loss}
\end{equation}
where $\mathcal D_{\mathrm{train}}$ is the controller-training distribution,
$\sigma$ is the logistic sigmoid, $\beta>0$ controls the pairwise preference
scale, $\lambda\geq0$ controls explicit controller regularization,
$W$ ranges over trainable controller tensors, and $|W|$ is the number of scalar
elements of $W$.

All base-model parameters $\omega$ remain frozen. Only controller parameters
$\theta$ receive gradients. We use
\begin{equation}
\lambda=10^{-4},
\label{eq:app_lambda}
\end{equation}
in addition to the decoupled weight decay of AdamW~\citep{loshchilov2019adamw}. The explicit Frobenius
penalty is a parameter-space regularizer and is not interpreted as a policy-KL
penalty. No additional advantage threshold or explicit policy-KL term
is used in this training objective.

The task-specific response score is
\begin{equation}
s_{e,\theta}
=
\begin{cases}
s_{\theta}^{\mathrm{margin}}, & \text{ConFiQA},\\
s_{\theta}^{\mathrm{logp}}, & \text{IMDb},\\
s_{\theta}^{\mathrm{logp}}, & \text{TL;DR}.
\end{cases}
\label{eq:app_task_training_score}
\end{equation}

Positive and negative RCM banks are trained toward the same target preference.
Their labels record the native RCM effects of the selected sites and do not
impose the sign of the learned perturbation.

The constant actuator contains one trainable vector
$v_i\in\mathbb{R}^{d_i}$ per controlled site. The rank-$k_i$ actuator contains
\begin{equation}
A_i\in\mathbb{R}^{k_i\times d_i},
\qquad
B_i\in\mathbb{R}^{d_i\times k_i},
\label{eq:app_lowrank_shapes}
\end{equation}
and therefore contributes $2d_i k_i$ trainable matrix parameters per site,
excluding any implementation-specific scalar gates. Reported parameter counts
are taken from the instantiated controller rather than inferred from nominal
rank alone.

Timing configurations are trained separately unless explicitly described as an
inference-only timing audit.

\subsection{Intervention Timing}
\label{app:timing}

The intervention mask determines which hidden states are modified; it is
separate from the response positions used to compute the training score.

\paragraph{\texttt{all}.}
During teacher-forced training, the \texttt{all} configuration permits direct
intervention at all registered positions in the complete prompt--response
sequence. During autoregressive generation, intervention remains enabled
in the initial prompt forward pass and the subsequent decoding passes. Each
decoding forward pass directly intervenes on the current decision state used
to produce the next-token distribution. Earlier controlled states affect later
computation through the ordinary autoregressive state and cache.

\paragraph{\texttt{prefill}.}
During training, the \texttt{prefill} mask applies only to prompt positions.
At inference time, it applies to all registered prompt positions during the
initial prompt forward pass and is then disabled for direct decode-time
intervention. The modified prompt computation and resulting cached states can
nevertheless continue to affect subsequent generation.

\paragraph{\texttt{prompt\_last}.}
When used as a diagnostic configuration, \texttt{prompt\_last} restricts direct
intervention to the final prompt position. It is therefore distinct from
\texttt{prefill}, which can modify multiple prompt positions.

The exact timing mode associated with every reported experiment is stored with
the corresponding run configuration. Timing variants are not silently pooled
as one condition.


\section{Additional RCM Analyses}
\label{app:rcm_analysis}

\subsection{RCM-Patch}
\label{app:rcm_patch}

RCM-Patch measures the effect of replacing one component state with a source state or a constructed prototype in the receiving computation; downstream states are recomputed.

Let $x^{\mathrm{ref}}$ denote the receiving input and
$x^{\mathrm{src}}$ the source condition. For structural component $i$, let
$h_i^{\mathrm{ref}}$ and $h_i^{\mathrm{src}}$ denote the corresponding native
component states at the matched registered positions. Let
$M_i^{\mathrm{ref}}$ denote the unmodified receiving run and
$M_{i\leftarrow h_i^{\mathrm{src}}}^{\mathrm{ref}}$ the same run after replacing
only the registered state of component $i$. The Patch effect is
\begin{equation}
\Delta_{i,e}^{\mathrm{patch}}
(x^{\mathrm{ref}};x^{\mathrm{src}})
=
C_e
\!\left(
M_{i\leftarrow h_i^{\mathrm{src}}}^{\mathrm{ref}};
x^{\mathrm{ref}}
\right)
-
C_e
\!\left(
M_i^{\mathrm{ref}};
x^{\mathrm{ref}}
\right).
\label{eq:app_patch}
\end{equation}

The receiving prompt, preference definition, scoring rule, intervention
positions, and downstream model computation remain fixed. Only the registered
component state is replaced. All downstream states are recomputed after the
substitution.

A positive Patch effect therefore means that the source state increases the
target preference score relative to the reference state at the receiving
condition. It does \emph{not} mean that the component's native presence has a
positive effect; native presence is measured by RCM-Zero.

\paragraph{Prototype states.}
When an experiment uses a prototype rather than a single paired source state,
let $\mathcal S_e$ denote the source set for preference $e$. The prototype is
\begin{equation}
\bar h_{i,e}^{\mathrm{src}}
=
\frac{1}{|\mathcal S_e|}
\sum_{x'\in\mathcal S_e}
h_i(x'),
\label{eq:app_patch_prototype}
\end{equation}
after applying the experiment's fixed position-alignment rule. The prototype
replaces $h_i^{\mathrm{src}}$ in Equation~\ref{eq:app_patch}. The reported
Site RCM-Patch results use prototypes, not a newly sampled paired source for
each receiving input. ConFiQA averages target-answer trajectories from the
context-supported condition and patches the receiving RAG prompt; IMDb
averages positive-prompt trajectories and patches the neutral receiving prompt.
The TL;DR prototype averages decision states within each preferred summary,
then across aliases and source inputs. In each case the receiver and scorer
are otherwise unchanged.
Each available source continuation contributes the mean of its decision
states, rather than one vote per token; where a question has multiple target
aliases, their means are averaged before the equal-weight mean across source
inputs. Accumulation and storage use float32, followed by a cast to the
receiving component's output dtype. The resulting head vector is broadcast
to each registered receiving decision state and overwrites that one head.
Fixed-response contrasts score both teacher-forced endpoints; IMDb's
generated-score contrast follows its receiving generation trajectory. The
prototype is neither centered nor normalized.
Patch candidates are split by the sign of their own mean paired Patch effect
and ranked within each side by the same directional confidence-bound rule
as Equation~\ref{eq:app_directional_ci}. For the vector Site configurations, the four-per-side budget has no
confidence-interval admission threshold; ties use ascending layer and head
indices.

Patch effects are treated as state-substitution effects and are not interpreted
as a decomposition of the component's complete native responsibility.

\subsection{Bilateral Effects Across Preference Domains}
\label{app:thresholds}
\label{app:roles}

For the scanned component set $\mathcal I$ and an absolute effect threshold
$\tau$ in the task's own score units, bilateral coverage is
\begin{equation}
\widehat p_{\mathrm{comp}}(\tau)
=
\frac{1}{N}\sum_{n=1}^{N}
\mathbf{1}\!\left[
\max_{i\in\mathcal I}\Delta_{i,e}^{\mathrm{zero}}(x_n)>\tau
\;\wedge\;
\min_{i\in\mathcal I}\Delta_{i,e}^{\mathrm{zero}}(x_n)<-\tau
\right].
\label{eq:app_competition_rate}
\end{equation}
The two extrema are separate singleton interventions on the same unmodified
input, not a jointly ablated trajectory. The rate is conditional on the scanned
candidates and their granularity. The thresholds below are descriptive points
on the full sensitivity curves, not significance cutoffs or parameters selected
by the final test results.

\begin{figure}[!htbp]
\centering
\includegraphics[width=\linewidth]{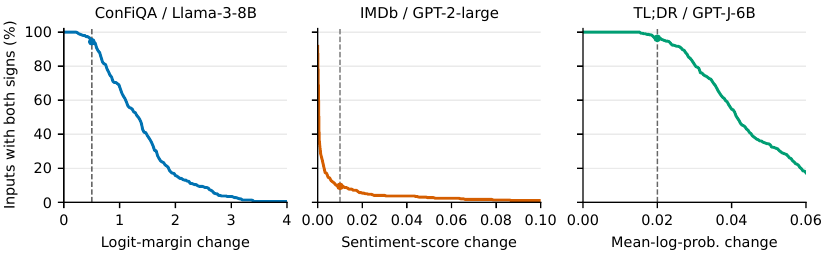}
\caption{\textbf{Cross-domain bilateral RCM-Zero coverage.}
All three scans use 300 discovery inputs. ConFiQA scans 32 attention-layer
writes and uses paired token-logit-margin changes; IMDb scans 160 attention
heads and uses generated sentiment-score changes; TL;DR scans 128 attention
heads and uses fixed-summary mean-log-probability changes. Each panel retains
its own score units and scanned set. Dashed lines mark the illustrative
thresholds in Table~\ref{tab:rcm_cross_domain}. The curves quantify signed
singleton effects, not simultaneous component ablations or the fraction of a
model's output explained by their sum.}
\label{fig:rcm_cross_domain}
\end{figure}

\begin{table}[!htbp]
\centering
\small
\begin{tabular}{lrrr}
\toprule
Task (scanned set) & $\tau$ & Both signs & Mean CI $+/-$ \\
\midrule
ConFiQA (32 layers) & 0.5 & 283/300 & 18 / 4 \\
IMDb (160 heads) & 0.01 & 28/300 & 52 / 0 \\
TL;DR (128 heads) & 0.02 & 289/300 & 3 / 4 \\
\bottomrule
\end{tabular}
\caption{\textbf{RCM-Zero effects on the frozen discovery inputs.}
``Mean CI $+/-$'' counts components whose pointwise paired-bootstrap interval
for the mean lies entirely above/below zero. These are not input counts on
which a component is active. IMDb effects are sparse: bilateral coverage falls
from 28/300 at $\tau=0.01$ to 3/300 at $\tau=0.1$.}
\label{tab:rcm_cross_domain}
\end{table}

Additional registered scans show the model dependence behind these
representative curves. On 300 ConFiQA inputs, the Qwen2-7B SFT scan finds both
signs above $\tau=0.5$ in 244/300 inputs across 224 scanned heads. For TL;DR
at $\tau=0.02$, the 128-head GPT-J DPO scan finds 299/300, while the 256-head
LION-LLaMA-3-8B scans find 300/300 before and after DPO. These are within-model,
within-score threshold comparisons; the numbers of scanned heads and score
units are not interchangeable across tasks. On the same per-component
criterion used below, 1/224 Qwen2-7B, 89/128 GPT-J DPO, and 120/256 and 189/256
LION SFT/DPO heads have effects exceeding the task threshold in both directions
on at least 10\% of inputs. The paired TL;DR scans therefore show persistent
bilateral effects after DPO without requiring the same head to retain the
same role.

In IMDb, 52 scanned heads have mean-effect intervals entirely above zero and
none has an interval entirely below zero. This weak competing-side signal
motivates the supporting-only deployment. The Site and Performance selection
rules are specified separately in Appendix~\ref{app:selection}.

\paragraph{DPO effect-change aggregation.}
The 1.00 margin-unit absolute change in Section~\ref{sec:residual_competition_results}
is $\frac{1}{N|\mathcal I|}\sum_{n=1}^{N}\sum_{i\in\mathcal I}
|\Delta_i^{\mathrm{DPO}}(x_n)-\Delta_i^{\mathrm{SFT}}(x_n)|$,
with $N=300$ paired inputs and $|\mathcal I|=32$ layers. Its 95\% interval
resamples the 300 paired inputs with replacement, retaining the 32-layer
vector for each sampled input (500 percentile-bootstrap resamples, seed 42).
It is not the absolute change of each layer's sample-mean effect.

\paragraph{Preference-relative component roles.}
The sign of a component's mean singleton effect describes its average
target-supporting or target-competing role under the specified task and
intervention. To distinguish a near-zero mean from genuinely small per-input
effects, we also report its mean absolute effect $\nu_i$ and
$\rho_i^+(\tau)=\mathbb P_x(\Delta_i(x)>\tau)$ and
$\rho_i^-(\tau)=\mathbb P_x(\Delta_i(x)<-\tau)$, using the empirical
frequencies over the same discovery inputs. For example, ConFiQA layer 14 has
mean $+0.072$ but mean absolute effect $1.090$ margin units, with effects above
$+0.5$ on 40\% of inputs and below $-0.5$ on 29\%. IMDb head L12H14 has mean
$+0.079$, mean absolute effect $0.453$, and respective frequencies 18.3\% and
13.7\% at $\tau=0.01$; TL;DR head L10H7 has mean $+0.0015$, mean absolute
effect $0.0283$, and frequencies 27.3\% and 24.7\% at $\tau=0.02$.
Across the scanned sets, 23/32, 14/160, and 27/128 components respectively
exceed their task's stated threshold in both directions on at least 10\% of
inputs. These descriptive roles are not a mutually exclusive causal partition
and do not imply additive effects.

\subsection{Target Outputs with Opposing Singleton Effects}
\label{app:target_outputs}

Two frozen ConFiQA QA cases connect the score contrast to the model's actual
native answer. In sample 000046, the question is ``Who owns Arsenal F.C.?'';
the supplied context supports ``Kering'' rather than the prior answer
``Kroenke Sports \& Entertainment.'' The native Llama-3-8B response is
``Kering.'' On this same input, attention-layer L30 has a $+1.758$ and L1 a
$-6.805$ native-minus-zero effect on the target--non-target logit margin.
In sample 000086, the supplied context supports ``J. K. Rowling'' rather than
``Tony McNamara'' for the creator of \emph{The Great}; the native response is
``J. K. Rowling.'' L0 has a $+4.340$ effect and L18 a $-3.406$ effect. Each
value comes from a separate single-component ablation with the fixed answer
endpoints. Thus a generated target answer can coexist with large opposing
native effects. These fixed-endpoint score contrasts do not themselves
establish that the ablated model would decode a different answer.

\subsection{Decoded Answer Changes under Single-Head Ablation}
\label{app:singleton_cases}

We separately checked whether a single-head ablation changes the decoded
answer, rather than only a fixed-answer score. On frozen ConFiQA QA test
inputs, we compared native Llama-3-8B-Instruct generation with generation under
RCM-Zero of pre-output-projection head \texttt{L17.attn.h5}. The head has a
negative mean native-minus-zero RCM effect of $-0.39176$ on the registered
target--non-target margin. Both runs use the same prompt and greedy decoding
with at most 64 new tokens and EOS enabled. The context-supported answer is
the task target, not necessarily the real-world answer.

The check covered 866 test inputs in fixed order. String-based scoring flagged
36 candidates; the following ten were selected after reading both complete
outputs and verifying that the ablated answer affirms the context-supported
target whereas the native answer does not. Because these are selected cases,
neither 36/866 nor 10/866 estimates a semantic flip rate. The outputs below
are decoded completions, not teacher-forced answers or margin predictions.

\begin{enumerate}
\item \textbf{Test 232: Edgar Allan Poe; target: dance music.}
Native: ``Literary fiction, specifically Gothic fiction and Dark Romanticism.''
Zero: ``According to the passage, the genre associated with Edgar Allan Poe is dance music.''

\item \textbf{Test 299: Pharrell Williams; target: Michelle Grace.}
Native: ``Michelle Larcher de Brito (n\'ee Williams), a fashion designer and artist.''
Zero: ``Pharrell Williams' spouse is Michelle Grace, a fashion designer and artist.''

\item \textbf{Test 422: Planet of the Apes (2001); target: Tim Story.}
Native: ``Tim Story did not direct the 2001 film `Planet of the Apes'. The actual director of the 2001 film `Planet of the Apes' is Tim Burton.''
Zero: ``According to the text, the director of the Planet of the Apes (2001 film) is Tim Story.''

\item \textbf{Test 470: Medici (TV series); target: superhero film.}
Native: ``The Medici (TV series) belongs to the genre of Historical Drama.''
Zero: ``According to the text, the Medici: Masters of Florence series belongs to the genre of `superhero film'.''

\item \textbf{Test 554: Donald Glover; target: action film.}
Native: ``According to the passage, Donald Glover represents the genres of comedy, acting, and rap.''
Zero: ``According to the passage, Donald Glover represents the genre of action film, as well as music (specifically, rap).''

\item \textbf{Test 636: Sea of Thieves; target: action thriller.}
Native: ``Action-adventure''.
Zero: ``According to the text, the genre of Sea of Thieves is an action thriller.''

\item \textbf{Test 660: The Conjuring: The Devil Made Me Do It; target: David Leitch.}
Native: ``Michael Chaves, not David Leitch.''
Zero: ``According to the text, the director of The Conjuring: The Devil Made Me Do It is David Leitch.''

\item \textbf{Test 697: Axel Witsel; target: Joe Biden.}
Native: ``Axel Witsel is a Belgian professional footballer, not American. He was born in Li\`ege, Belgium, not the United States of America. Therefore, the head of state of the country where Axel Witsel holds citizenship is the King of Belgium, currently Philippe.''
Zero: ``According to the text, Axel Witsel is a professional footballer who plays for the Turkish club Be\c{s}ikta\c{s} and the United States of America national team. Therefore, the head of state of the country where Axel Witsel holds citizenship is President Joe Biden, who is the President of the United States of America.''

\item \textbf{Test 713: Sridevi; target: United States dollar.}
Native: ``The question is a bit tricky because the text states that Sridevi is a renowned American actress, but it also mentions that she was born in Sivakasi, Tamil Nadu, which is a location in India, not the United States. Therefore, it is not possible to determine the currency of the country where S''
Zero: ``The answer is USD, which is the United States dollar.''

\item \textbf{Test 833: Aymeric Laporte; target: North America.}
Native: ``The answer is Europe. Aymeric Laporte was born in Narbonne, France, not the United States of America. He has represented France at various youth levels and made his senior international debut for France in 2016, not the United States of America.''
Zero: ``The answer is North America.''
\end{enumerate}

In the Tim Story and David Leitch examples, the native output mentions the
target while explicitly denying it. A lexical answer-occurrence score can
therefore mark both native and ablated texts as target hits; the semantic flip
in these examples follows from the complete generated statements. The native
completion for test 713 ends mid-sentence at the 64-token limit.


\section{Experimental and Implementation Details}
\label{app:implementation}

\subsection{Optimization Details}
\label{app:optimization}

All CAST runs freeze the base language-model parameters and update only the
controller parameters. AdamW~\citep{loshchilov2019adamw} is used for controller
optimization. The explicit controller penalty uses $\lambda=10^{-4}$ as
defined in Equation~\ref{eq:app_cast_loss}; any decoupled AdamW weight decay
is reported separately and must not be described as policy-KL regularization.

Table~\ref{tab:cast_hparams} reports the locked SFT-starting CAST settings.
Within a task, the two reported model families use the same pair counts and
optimizer settings; \texttt{all} and \texttt{prefill} are separately trained.

\begin{table}[!htbp]
\centering
\small
\begin{tabular}{lccc}
\toprule
Setting & ConFiQA & IMDb & TL;DR \\
\midrule
Training pairs & 240 & 1,024 & 8,192 \\
Validation pairs & 60 & 256 & 512 \\
Epochs & 2 & 1 & 1 \\
Learning rate & 0.05 & 0.05 & 0.001 \\
Training batch & 4 & 32 & 2 \\
$\beta$ & 1.0 & 1.0 & 0.5 \\
AdamW weight decay & 0.01 & 0.01 & 0.01 \\
Random seed & 42 & 42 & 42 \\
\bottomrule
\end{tabular}
\caption{SFT-starting CAST optimization settings. ConFiQA and IMDb train the
two banks separately; the supporting-only IMDb configuration deploys
only the supporting bank. TL;DR trains 16 candidate single-head low-rank
controllers before its held-out head audit. Separate timing modes are treated
as independent training runs; inference-only timing audits are explicitly
identified as such.}
\label{tab:cast_hparams}
\end{table}

\begin{table}[!htbp]
\centering
\small
\begin{tabular}{@{}p{0.29\linewidth}p{\dimexpr0.71\linewidth-2\tabcolsep\relax}@{}}
\toprule
Execution choice & Frozen rule \\
\midrule
SV initialization & Each head vector $v_i$ starts at zero in float32. \\
Low-rank initialization & $A_i$ is drawn from a zero-mean normal with standard deviation $0.01$ and $B_i$ starts at zero, both in float32. The TL;DR run draws the head-by-rank layout before transposing $A_i$; the initial intervention is zero. \\
Training strength and checkpoint & $\alpha_{\rm train}=1$ throughout optimization. The payload saved after the configured final epoch is used; validation does not select an intermediate checkpoint. \\
DPO start & Transfer (T) applies the saved SFT-start payload to the released DPO checkpoint without further controller training. Nontransfer (NT) initializes a new controller and trains it on the DPO checkpoint with the same task-level optimization values in Table~\ref{tab:cast_hparams}. Each row retains its registered training and inference timing. \\
Bank deployment & The retained banks are summed without normalization or joint retraining and multiplied by one shared inference strength $\alpha$, not independently calibrated strengths. \\
\bottomrule
\end{tabular}
\caption{Controller initialization, training, and deployment settings.}
\label{tab:cast_execution}
\end{table}

ConFiQA selects one shared $\alpha$ from $\{0.1,0.2,\ldots,1.0\}$ by
maximizing the non-test score
$r_{\rm context\,only}+r_{\rm short\,exact\,context}
-r_{\rm prior\,hit}-r_{\rm neither}-\bar\ell_{\rm chars}/400$;
exact ties choose the smallest strength. Here context-only means a
context-answer hit without a prior-answer hit; short-exact-context means
an exact context answer in at most 48 decoded characters; prior-hit includes
outputs mentioning both answers; neither means no answer hit; and
$\bar\ell_{\rm chars}$ is mean decoded length in characters. TL;DR searches
$\{0.2,0.3,\ldots,0.7\}$ and maximizes
the two-order balanced win rate against the human summary on its strength
calibration posts. Strengths within $0.003$ of the best rate are eligible,
with the smallest eligible $\alpha$ chosen. IMDb reports the full
$\{0.1,0.2,\ldots,1.0\}$ response curve rather than selecting one point by
the final test. Transfer recalibrates on the DPO inference checkpoint's
non-test data. Test-set outcomes are not used to select $\alpha$, component
sets, ranks, checkpoints, or other operating parameters.

\subsection{Model and Hardware Details}
\label{app:model_hardware}

The starting/DPO checkpoint pairs are Meta-Llama-3-8B-Instruct /
Context-Faithful-LLaMA-3-8B and Qwen2-7B-Instruct /
Context-Faithful-Qwen2-7B for ConFiQA; ma921 GPT-2-large SFT / DPO for IMDb;
and CarperAI GPT-J TL;DR SFT / quyanh GPT-J DPO and Columbia-NLP
LION-LLaMA-3-8B SFT / DPO for TL;DR. These DPO models are released
checkpoints, not retrainings on CAST's controller pairs. The
condition index supplies their full public model identifiers, immutable
revisions, tokenizer bindings, precision, and per-run software environment.
The accompanying materials are indexed by
\texttt{EVIDENCE\_INDEX.json} at
\url{https://github.com/abovefiramament/DHSA}.
The ConFiQA Qwen2-7B SFT+CAST-SV rows and transferred DPO+CAST-SV rows
use model revision
\texttt{f2826a00ceef68f0f2b946d945ecc0477ce4450c}.
The Performance comparisons used 80-GB H100 devices with baseline-specific
dependency profiles; the Site archive records its own device profile rather
than assuming the Performance environment applies to it.

The base model families are GPT-2~\citep{radford2019gpt2},
GPT-J~\citep{wang2021gptj}, Llama 3~\citep{grattafiori2024llama3},
Qwen2~\citep{yang2024qwen2}, and Qwen2.5~\citep{qwen2024qwen25}.
The Context-Faithful checkpoints originate from
Context-DPO~\citep{bi2025contextdpo}; the LION SFT/DPO checkpoints originate
from the LION alignment study~\citep{yu2024lions}.

\begin{table}[!htbp]
\centering
\small
\setlength{\tabcolsep}{3.5pt}
\begin{tabular}{llrrr}
\toprule
Task/model & CAST operator & Rank & Active heads & Learned scalars \\
\midrule
ConFiQA / Llama-3-8B & SV & -- & 8 & 1,024 \\
ConFiQA / Qwen2-7B & SV & -- & 8 & 1,024 \\
IMDb / GPT-2-large & SV, two banks & -- & 8 & 512 \\
IMDb / GPT-2-large & SV, support-only & -- & 4 & 256 \\
IMDb / GPT-2-large & low-rank, two banks & 4 & 8 & 4,096 \\
TL;DR / GPT-J-6B & low-rank & 4 & 8 & 16,384 \\
TL;DR / LION-LLaMA-3-8B & low-rank & 4 & 8 & 8,192 \\
\bottomrule
\end{tabular}
\caption{Actual active controller parameter counts from the frozen controller
tensors. Separate \texttt{all} and \texttt{prefill} runs have the same counts.
The main IMDb efficiency curve deploys the support-only bank; its four heads
are a subset of the eight-head trained candidate. TL;DR trains 16 candidate
single-head controllers and retains eight after audit, without joint
retraining. Candidate-training cost and final active size are reported separately.}
\label{tab:controller_parameters}
\end{table}

For head-level CAST, interventions are applied to the registered head slice
before the attention output projection. For layer- or MLP-level CAST, the
corresponding residual-write interface is used. The model implementation,
tokenizer, and all non-controller parameters remain unchanged within each
comparison.
IMDb controller pairs originate from \texttt{ma921/imdb-tokenized\_noise0}
at revision \texttt{5ad0e4fbb89a87404918f5f510c9192afb3d350a}; its
test comments originate from \texttt{stanfordnlp/imdb} at revision
\texttt{e6281661ce1c48d982bc483cf8a173c1bbeb5d31}. TL;DR pairs and
test-post projections derive from \texttt{openai/summarize\_from\_feedback}
at revision \texttt{b8f7d168b6f4e95b2a92e84768bd6c955bed2f29}.
The indexed per-condition configurations identify the exact public source
URLs, model revisions, and tokenizer binding; these pair sources must not
be inferred from a released DPO checkpoint's name alone.

\subsection{Baseline Configurations}
\label{app:baselines}

All methods in a given task/model comparison use the same final input set,
decoding rule, and evaluator. This is an evaluation-matched comparison; it is
not a claim that public DPO checkpoints or the external methods share CAST's
training pairs or the same search cost. Search and final active controller
size are separate quantities.

\paragraph{DPO.}
DPO is the main preference-optimization reference. Every DPO row in the main
Performance table evaluates a released checkpoint. Its original preference
data may differ from CAST's frozen pair set; only the final evaluation is
shared. DPO+CAST-T reuses an SFT-trained controller and recalibrates strength
on non-test data; DPO+CAST-NT trains a new controller on the DPO checkpoint.

\paragraph{LoReFT.}
LoReFT retains the published low-rank representation-intervention operator and
training objective~\citep{wu2024reft}. Its four-layer joint tuples and
strength grid are selected on non-test data. Mapping published layer depths
to a different architecture is our registered adaptation, not an author-provided
task default.

\paragraph{BiPO.}
BiPO retains its bidirectional preference objective and one learned steering
vector per candidate layer~\citep{cao2024bipo}. It is not converted to CAST's
signed-bank training. Its CAA-based layer preselection and candidate
calibration remain separate from RCM. We explicitly apply the registered
AdamW weight decay of 0.05; this is a project adaptation because the released
training script exposes that value without forwarding it to its DPO
configuration.

\begin{table}[!htbp]
\centering
\small
\setlength{\tabcolsep}{4pt}
\begin{tabular}{lrrrr}
\toprule
Task/model & CAST trained & CAST retained & BiPO layers & LoReFT tuples \\
\midrule
ConFiQA / Llama-3-8B & 8 & 8 & 6 & 8 \\
ConFiQA / Qwen2-7B & 8 & 8 & 6 & 8 \\
IMDb / GPT-2-large & 8 & 4--8 & 10 & 6 \\
TL;DR / GPT-J-6B & 16 & 8 & 7 & 1 \\
TL;DR / LION-LLaMA-3-8B & 16 & 8 & 14 & 2 \\
\bottomrule
\end{tabular}
\caption{Candidate-training opportunity by task/model, not a claim of equal
GPU cost. CAST columns count heads. Its TL;DR candidates are individually
trained heads; BiPO independently trains each registered layer, whereas each
LoReFT tuple is a joint four-layer controller. BiPO uses 20/5/1 epochs on
ConFiQA/IMDb/TL;DR; LoReFT uses 24 epochs, while CAST uses the epochs in
Table~\ref{tab:cast_hparams}. All candidate selection and strength calibration
precede final test.}
\label{tab:candidate_training}
\end{table}

For IMDb, the registered 80-GB H100 pre-test occupancy was approximately
2,448 seconds for SFT CAST-SV-all, 2,343 seconds for the expanded ten-layer
BiPO controller search, and 2,410 seconds for the six-tuple LoReFT search.
The boundaries are CAST's start-to-pretest pipeline versus external
controller jobs; final generation/evaluation is excluded. These durations
show comparable observed occupancy for this task, not matched FLOPs, and no
strict compute-equality claim is made for the other tasks.

\begin{table}[!htbp]
\centering
\small
\begin{tabular}{lp{0.28\linewidth}p{0.43\linewidth}}
\toprule
Task & RCM search & Final selection \\
\midrule
ConFiQA & Four parent layers per signed side; every head in their union & Four heads per side by each head's own directional confidence bound; no post-training head audit. \\
IMDb Site & Four parent layers per signed side; every head in their union & Four positive- and four negative-effect heads by directional confidence bound. \\
IMDb Performance & Four parent layers at each high/low effect extreme; every head in their union & Four heads per extreme by CI bound, regardless of the low-end sign; support-only activates the high-effect bank. \\
TL;DR & Four parent layers per side; every head in their union & Eight signed candidates per side are trained individually; a 40-post non-test audit retains eight without a final sign quota. \\
\bottomrule
\end{tabular}
\caption{Task-specific selection and audit budgets. Each condition's ordered
head identifiers, bank composition, and final strength are recorded in the
position, controller, and strength manifests linked by the artifact index.}
\label{tab:selection_execution}
\end{table}

The TL;DR audit tests each trained single-head payload against the human
reference at $\alpha\in\{0.2,0.3,\ldots,0.7\}$ in both display orders.
Its review packet replaces head coordinates with audit IDs and includes
balanced win rate, output-health measures, the post, human summary, and
generated text. Human review retains a clean gain with healthy summaries;
it can also retain a candidate whose high-strength overshoot is recoverable
at a lower strength. Persistent factual errors, repetition, unfinished
outputs, and weaker or redundant benefit count against retention under the
fixed eight-head budget. These are qualitative keep/drop principles, not a
scalar sorting rule or numerical tie-break. The decisions are fixed before
final test. Retained tensors are
united directly without normalization, rescaling, sign rebalance, or an
optimizer step. The GPT-J SFT-all freeze contains two target-support and
six competitor-support heads, confirming that no four/four final quota is
imposed. In the TL;DR Site comparison, RCM-Zero, RCM-Patch, ITI, and each
Random seed each receive 16 single-head training opportunities and the same
40-post review, retaining eight heads without joint retraining. TL;DR ITI
labels preferred and rejected summaries as positive and negative states,
uses the final state of each teacher-forced continuation, and ranks heads by
grouped two-fold probe accuracy before this common audit. Random candidates
are drawn without replacement from architecture-compatible heads.

\paragraph{Selector controls.}
In the ConFiQA and IMDb vector Site comparisons, RCM constructs two signed
four-head candidate banks and trains them separately toward the same target
preference. The supporting-only IMDb configuration leaves the weak competing
bank inactive at deployment without changing the candidate rule. ITI and the
three fixed random selectors each use one jointly trained unsigned eight-head
bank. These comparisons measure the registered selector-plus-bank
configurations, not the site-ranking rule in isolation. TL;DR uses its
separate single-head audit and eight-head composition protocol, so the
vector two-bank description does not apply there. Final inputs and evaluators
are shared within each comparison. Random sites are architecture-compatible
and retain their fixed seeds.
ITI trains one L2-regularized logistic probe per pre-output attention head
on the same paired discovery inputs as RCM. It reads the final non-padding
state of each complete labeled continuation and ranks heads by mean
held-out accuracy in two grouped folds (train A/test B, then reverse).
Both endpoints of a pair stay in the same fold; accuracy ties use ascending
layer and head indices. ConFiQA labels identify canonical
context-supported versus prior-answer continuations. IMDb labels identify
positive-prompt versus neutral-prompt generations with the shared native
token sequences; neither raw IMDb sentiment labels nor reward-ranked pairs
are substituted. ITI's eight selected heads form one unsigned bank.

\subsection{Data Separation and Selection Hygiene}
\label{app:data_splits}

We distinguish RCM selection, controller training, validation, strength/audit
calibration where applicable, and final test by their uses. The final test
is held out from selection, training, and calibration; other roles are not
universally disjoint. In ConFiQA, controller training and validation are
partitioned from the same 300 admitted groups used by the selector. In the
registered TL;DR procedure, selector inputs may overlap controller-training
inputs, while audit, strength-calibration, and final-test posts are separately
assigned. The principal Performance rows use respectively
300/240/60/120/2,048 inputs or pairs for ConFiQA
(selector/train/validation/strength/test), 300/1,024/256/2,048 for IMDb
(selector/train/validation/test), and 300/8,192/512/40/40/512 for TL;DR
(selector/train/validation/head audit/strength/test). ConFiQA's final 2,048
inputs mix QA, MR and MC; the TL;DR final set has 512 posts. IMDb's 1,024
training pairs retain the public token-pair source after its frozen cleaning
and split rule; its 300 selector prompts and 2,048 official test comments
are distinct from the training pool. The TL;DR 16-head audit uses 40 non-test
posts and evaluates each candidate against the human reference in both
presentation orders; the selected eight-head combination is not jointly
retrained. Calibration data are small relative to training, but candidate
audit and strength search are included in the method's acquisition cost
rather than hidden inside the final eight-head parameter count.

No final-test example selects a component, controller checkpoint, rank, head
combination, or strength. The condition index contains the exact
row identifiers, final sites, strengths, checkpoints, scoring traces, public
source revisions, and execution environments for each row; the appendix
reports the scientifically relevant comparisons rather than duplicating
that machine-readable inventory.

\subsection{Final Evaluation Definitions}
\label{app:evaluation}

\paragraph{ConFiQA.}
The scorer case-folds generated text and answer aliases, replaces punctuation
by spaces, and collapses whitespace. PC is one if any context-supported answer
alias occurs as a whitespace-bounded phrase anywhere in the output. The
analogous prior-answer hit is PO. EM is one only if the entire normalized
output equals a context-supported alias. Thus PC need not imply an exact or
well-formed answer. For example, in frozen QA row 001894 the target alias is
``Jane Conrad'' and CAST-all outputs ``Jane Conrad, a social worker and
activist.'' This has PC $=1$ and EM $=0$. Site reports the macro-average PC
across QA, MR and MC; Performance reports the combined 2,048-input test
result. PC is an occurrence metric, so a negated mention of a target answer
can still count as a hit; the decoded examples in
Appendix~\ref{app:singleton_cases} are also checked for whether the model
affirms that answer.

\paragraph{IMDb.}
The frozen evaluator is SiEBERT~\citep{hartmann2023siebert}, using
\texttt{siebert/sentiment-roberta-large-english} at revision
\texttt{74cea614e245}. Its reported sentiment reward is
$P(\mathrm{POSITIVE})-P(\mathrm{NEGATIVE})$, not an unbounded learned reward;
positive rate is the fraction with reward greater than zero. Both
distribution-shift axes compare the evaluated policy to the registered SFT
reference on its generated continuation $y$:
\begin{equation}
\begin{split}
K_{\mathrm{seq}}(x,y)
&=\sum_t\log\frac{p_M(y_t\mid x,y_{<t})}{p_{\mathrm{SFT}}(y_t\mid x,y_{<t})},\\
K_{\mathrm{tok}}(x,y)
&=\frac{1}{|y|}\sum_t
D_{\mathrm{KL}}\!\left(p_M(\cdot\mid x,y_{<t})\,\|\,
 p_{\mathrm{SFT}}(\cdot\mid x,y_{<t})\right).
\end{split}
\label{eq:imdb_distribution_shift}
\end{equation}
The main GPT-2-large Site plot uses the average $K_{\mathrm{seq}}$, labeled
\emph{sequence log-ratio}; the Performance plot uses the average
$K_{\mathrm{tok}}$. $K_{\mathrm{seq}}$ is a generated-sequence log-probability
ratio, not the tokenwise full-vocabulary KL. Under deterministic decoding,
it is not an unbiased Monte Carlo estimate of policy sequence KL. The
additional Qwen2.5-14B Site plot uses full-vocabulary token KL
(Appendix~\ref{app:imdb_site}). Generation health is the nonempty-output
rate plus mean completion length. At the plotted DPO point these are
99.90\% and 230 tokens; for support-only CAST-SV-all at $\alpha=0.5$ and
$0.6$ they are 100\%/215 tokens and 99.95\%/212 tokens, respectively.
Reward and KL are always reported together, so a higher reward obtained only
with much greater policy shift is not treated as an unqualified improvement.
All 2,048 IMDb test inputs remain in the sentiment-reward, KL, and nonempty-rate
denominators, including decoded empty strings; the empty string is sent to the
same SiEBERT classifier. In the GPT-2 Performance records, a sole generated EOS
can decode to empty text but remains one generated token and one token-KL step.
For a genuinely zero-token continuation, the Site evaluator records
$K_{\mathrm{tok}}=0$ rather than dividing by zero; this occurs in the reported
Qwen2.5-14B Site curve, whose reward and KL averages retain all 2,048 inputs.

\paragraph{TL;DR.}
The pairwise evaluator is the registered \texttt{deepseek-v4-pro} API alias;
the provider does not publish an immutable revision for that alias, so the
saved request and response traces are the version evidence. The system
message is ``You are an impartial evaluator. Follow the requested comparison
criteria exactly. The summaries are anonymous. Return only a valid JSON
object.'' The user message is:
\begin{quote}
Which of the following summaries does a better job of summarizing the most
important points in the given forum post, without including unimportant or
irrelevant details? A good summary is both precise and concise.

Post: [post]\\
Summary A: [summary A]\\
Summary B: [summary B]

First provide a one-sentence comparison of the two summaries, explaining
which you prefer and why. Then choose only A or B. Return JSON exactly as:
\texttt{\{"comparison": "one-sentence comparison and explanation", "preferred": "A or B"\}}
\end{quote}
The first A/B order is deterministic from the sample ID and seed; the second
call swaps the order. A row scores
$(\mathbf{1}[\text{candidate wins first}]+\mathbf{1}[\text{candidate wins swap}])/2$,
so its value is 0, 0.5, or 1; the table reports the mean over rows. There is no
tie label or imputed judgment: a malformed response is retried up to five
times and then fails evaluation. Empty model outputs remain in the comparison
rather than being silently dropped. All methods use the same human references,
prompt and two-order rule. The reported win rates are not head-to-head
comparisons between CAST and DPO.
The saved API configuration uses \texttt{deepseek-v4-pro} with thinking
disabled, \texttt{reasoning\_effort=high}, an 800-token response cap,
a 180-second timeout, and at most five retries. Each of the 20 TL;DR
Performance conditions has 512 posts with both presentation orders judged.
No post exhausted retries, so every reported win rate uses all 512 posts.
The provider does not publish an immutable revision for this model alias;
the saved requests and responses document the calls.

\begin{table}[!htbp]
\centering
\small
\begin{tabular}{lp{0.27\linewidth}p{0.40\linewidth}}
\toprule
Task & Final prompt and input handling & Decoding and stopping \\
\midrule
ConFiQA & \texttt{context, Q: question, A:} prompt, wrapped in the model's registered chat template for instruct checkpoints. No additional input-token cap or truncation is set by the final-generation callback. & Greedy; at most 64 new tokens; registered \texttt{Q:} stop string and native EOS; seed 42. \\
IMDb & A tokenizer-derived 2--8-token review prefix, without a chat template. No final-generation input truncation is set. & One sampled continuation per prompt; at most 256 new tokens; temperature 1, top-$k$ 50, top-$p$ 1, seed 42, native EOS. The same random-number schedule is reused across strengths. \\
TL;DR & The prepared source post prompt is passed through without an additional chat template or final-generation truncation. & Greedy; at most 100 new tokens; seed 42, native EOS, model KV cache enabled. The registered top-$p=0.9$ and top-$k=0$ do not affect greedy selection. \\
\bottomrule
\end{tabular}
\caption{Final-generation settings. ``No input cap'' means no extra truncation in this evaluation
path; the model's context capacity still applies. RCM discovery generation
has its own registered settings and is not inferred from this table.}
\label{tab:final_generation}
\end{table}

\subsection{Paired Uncertainty for Main Performance Contrasts}
\label{app:performance_ci}

Table~\ref{tab:paired_main_effects} reports selected main-table differences
using the saved per-input scores. For each comparison, the same test inputs
are paired by ID; we resample inputs with replacement 5,000 times using
seed 42 and take the 2.5th and 97.5th percentiles of the mean paired
difference. TL;DR's first and swapped presentation-order judgments are first
averaged within each post, so the 512 posts, not 1,024 judge calls, are the
resampling units.

\begin{table}[!htbp]
\centering
\small
\setlength{\tabcolsep}{5pt}
\begin{tabular}{llr}
\toprule
Task / metric & Contrast & Difference [95\% interval], pp \\
\midrule
ConFiQA Llama / PC & CAST-all $-$ DPO & $-6.98\;[-8.74,-5.27]$ \\
 & DPO+CAST-T-all $-$ DPO & $0.00\;[-0.88,0.88]$ \\
ConFiQA Llama / EM & CAST-all $-$ DPO & $-43.99\;[-46.19,-41.89]$ \\
 & CAST-prefill $-$ DPO & $20.65\;[18.50,22.80]$ \\
 & DPO+CAST-T-all $-$ DPO & $3.96\;[2.25,5.66]$ \\
ConFiQA Qwen / PC & CAST-all $-$ DPO & $3.66\;[2.05,5.27]$ \\
 & DPO+CAST-T-all $-$ DPO & $10.01\;[8.35,11.62]$ \\
ConFiQA Qwen / EM & CAST-all $-$ DPO & $12.45\;[10.84,14.01]$ \\
 & DPO+CAST-T-all $-$ DPO & $35.99\;[33.89,38.13]$ \\
TL;DR GPT-J / Win & CAST-all $-$ DPO & $-0.39\;[-3.91,3.12]$ \\
 & DPO+CAST-T-all $-$ DPO & $6.45\;[3.52,9.47]$ \\
TL;DR LION / Win & CAST-all $-$ DPO & $-5.57\;[-8.30,-2.93]$ \\
 & CAST-prefill $-$ DPO & $-1.86\;[-4.69,1.07]$ \\
 & DPO+CAST-T-all $-$ DPO & $0.88\;[-1.17,2.93]$ \\
 & DPO+CAST-T-prefill $-$ DPO & $2.34\;[-0.10,4.79]$ \\
\bottomrule
\end{tabular}
\caption{Paired percentile-bootstrap intervals for the Performance table,
in percentage points. ConFiQA uses 2,048 inputs per model; TL;DR uses 512
posts per model. The reproducible calculation is supplied as
\texttt{reproduction/paired\_main\_effects.py} with its JSON output in the
accompanying materials. These are fixed-checkpoint, fixed-test-set intervals,
not intervals over training seeds; an interval spanning zero does not
establish equivalence.}
\label{tab:paired_main_effects}
\end{table}

\subsection{Additional IMDb Site Result}
\label{app:imdb_site}

\begin{figure}[!htbp]
\centering
\includegraphics[width=\linewidth]{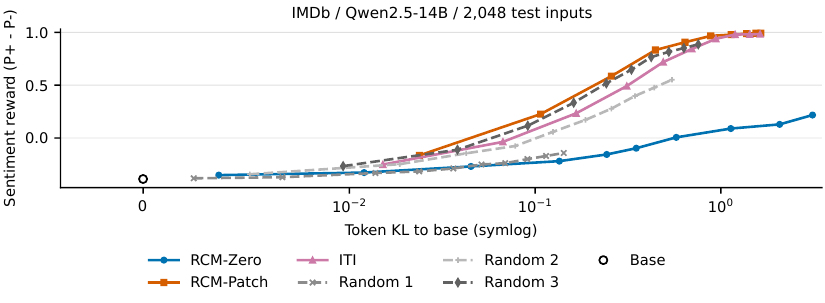}
\caption{\textbf{IMDb Site response curves for Qwen2.5-14B.}
Mean bounded sentiment reward against exact token KL to the same model without
control. The open circle is the unmodified base model. Each selector is
evaluated at ten registered strengths on 2,048 shared test prompts. Random
1--3 are fixed, independently reported seeds; no seed is selected by test score.}
\label{fig:imdb_qwen_site}
\end{figure}

Figure~\ref{fig:imdb_qwen_site} reports the second IMDb Site model family
omitted from the main efficiency panel. On Qwen2.5-14B, RCM-Patch, ITI and
one random seed all achieve large sentiment changes as strength increases,
whereas RCM-Zero is weaker over the recorded response curve. Thus the
GPT-2-large ranking is not assumed to transfer unchanged to Qwen. Every curve
contains all ten registered strengths on the same 2,048 test prompts; the
three random seeds remain separate. The archived Site score is mean
positive-class probability $p_+$ from a binary classifier, converted to the
paper's bounded reward $2p_+-1$. The horizontal axis is the full-vocabulary
token KL to the unmodified Qwen reference; it is not the sampled sequence
log-probability ratio used in the main GPT-2-large Site panel. Nonempty-output
checks remain relevant at large strengths: RCM-Zero reaches an empty-output
rate of 3.6\% at one grid point, while the other plotted selectors remain
at zero.

\subsection{IMDb DPO-Start Response Curves}
\label{app:imdb_dpo}

\begin{figure}[!htbp]
\centering
\includegraphics[width=\linewidth]{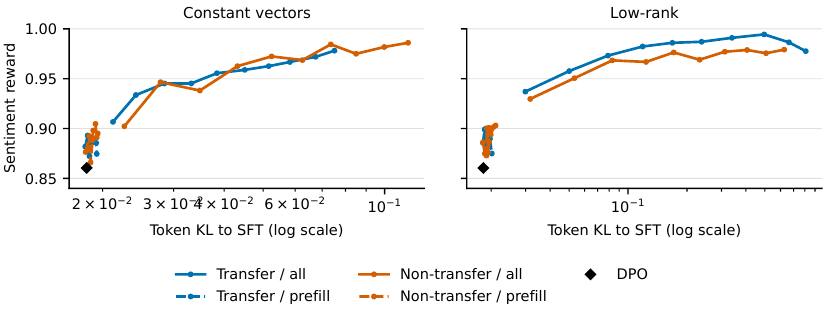}
\caption{\textbf{IMDb DPO+CAST, 2,048 frozen test prompts.}
The black diamond is the released DPO checkpoint. Each line contains the ten
registered controller strengths and reports the same bounded sentiment
reward and tokenwise KL to the SFT reference as the main IMDb plot.
Blue/orange denote SFT-to-DPO transfer/new DPO training; solid/dashed denote
all-token/prefill application. These are test response curves, not
test-selected optima.}
\label{fig:imdb_dpo_results}
\end{figure}

Figure~\ref{fig:imdb_dpo_results} completes the DPO+CAST comparison for IMDb
without listing every test strength. Transfer reuses the SFT-trained
controller; non-transfer trains on the DPO checkpoint. Both constant-vector
and low-rank controllers are shown at \texttt{all} and \texttt{prefill}
timings. The complete curves reveal the reward--distribution-shift tradeoff:
\texttt{all} can raise sentiment reward substantially, whereas the
\texttt{prefill} changes are smaller and stay close to the DPO point.
The plotted strength grid is descriptive; no test point was used to choose
the deployed strength.

\begin{figure}[!htbp]
\centering
\includegraphics[width=\linewidth]{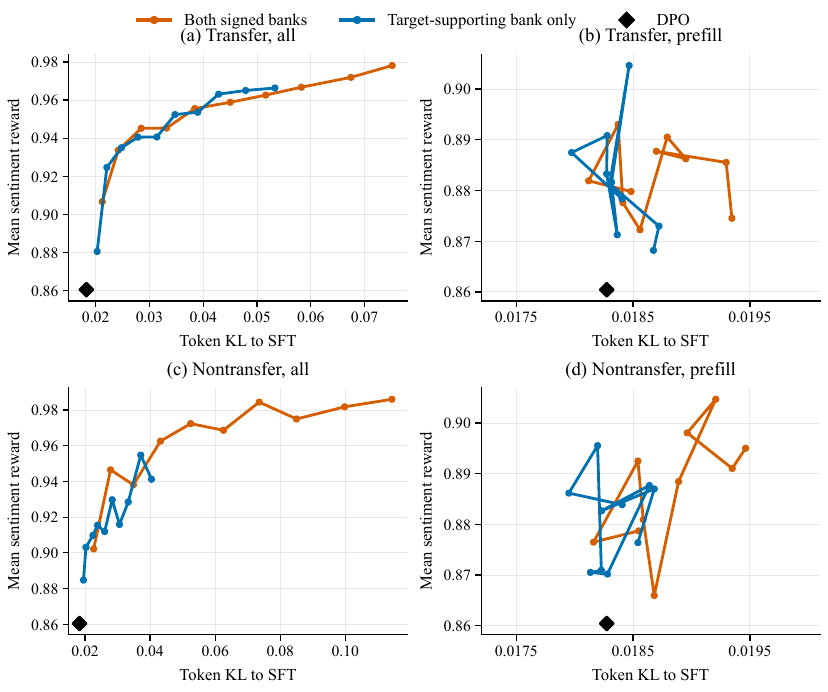}
\caption{\textbf{IMDb DPO+CAST two-bank ablation on H100.}
The original two-bank constant-vector controller (orange) is compared
with a replay using only its target-supporting four-head bank (blue).
Panels separate SFT-to-DPO transfer and DPO-start training, each under
\texttt{all} and \texttt{prefill} timing. Every curve uses the same
$\alpha\in\{0.1,0.2,\ldots,1.0\}$ grid on 2,048 fixed test prompts;
the black diamond is DPO. Vertical axes show bounded sentiment reward;
horizontal axes show token KL to the SFT reference. The grid describes
response curves and is not a test-set selection rule.}
\label{fig:imdb_dpo_bank_ablation}
\end{figure}

\paragraph{DPO-start two-bank ablation.}
Figure~\ref{fig:imdb_dpo_bank_ablation} reports the \texttt{all} and
\texttt{prefill} bank comparisons. The ablation replays the four frozen
constant-vector DPO+CAST controllers with only their four
target-supporting heads active. The original two-bank curves are the
matched controls: the DPO checkpoint, 2,048 test prompts, ten strengths,
generation settings, and evaluator are unchanged. No head is reselected
and no controller is retrained for this ablation. At $\alpha=0.4$ under
\texttt{all} timing, transfer yields reward/KL $0.9405/0.0279$ with the
supporting bank and $0.9453/0.0332$ with both banks; DPO-start training
yields $0.9155/0.0238$ and $0.9626/0.0432$, respectively. The
uncontrolled DPO point is $0.8604/0.0183$. Thus the supporting bank
alone can raise sentiment reward above DPO, while the full controller
can yield additional reward at greater distribution shift. The
\texttt{prefill} curves stay much closer to DPO. These observations do
not establish which bank configuration is superior at matched KL.

\section{Theoretical Details}
\label{app:theory}

\subsection{DPO-Form Interpretation of the Pairwise Objective}
\label{app:advantage_equivalence}

This section establishes the DPO-form representation of the pairwise
objective used in Section~\ref{sec:learning}, through a score-induced policy.

Fix an input $x$. Let $\mathcal Y_x$ denote the response space considered by
the protocol and let $\pi_0(y\mid x)>0$ be the frozen reference distribution on
that space. Define
\begin{equation}
u_{e,\theta}(x,y)
=
s_{e,\theta}(x,y)
-
s_{e,0}(x,y),
\label{eq:app_score_gain}
\end{equation}
where $s_{e,\theta}$ and $s_{e,0}$ are the controlled and reference response
scores.

Assume the normalizer
\begin{equation}
Z_{e,\theta}(x)
=
\sum_{y\in\mathcal Y_x}
\pi_0(y\mid x)
\exp\!\left(u_{e,\theta}(x,y)\right)
\label{eq:app_score_normalizer}
\end{equation}
is finite. Define the score-induced policy
\begin{equation}
\bar\pi_{e,\theta}(y\mid x)
=
\frac{
\pi_0(y\mid x)
\exp\!\left(u_{e,\theta}(x,y)\right)
}{
Z_{e,\theta}(x)
}.
\label{eq:app_score_policy}
\end{equation}

\paragraph{Proposition 1 (DPO-form representation).}
For every response pair $(y^+,y^-)$,
\begin{equation}
u_{e,\theta}(x,y^+)
-
u_{e,\theta}(x,y^-)
=
\log
\frac{\bar\pi_{e,\theta}(y^+\mid x)}
     {\pi_0(y^+\mid x)}
-
\log
\frac{\bar\pi_{e,\theta}(y^-\mid x)}
     {\pi_0(y^-\mid x)}.
\label{eq:app_dpo_form}
\end{equation}
Consequently, the pairwise term
\begin{equation}
-\log
\sigma
\!\left(
\beta
[
u_{e,\theta}(x,y^+)
-
u_{e,\theta}(x,y^-)
]
\right)
\label{eq:app_pair_loss}
\end{equation}
is exactly the reference-relative logistic form used by DPO
~\citep{rafailov2023} when applied to the induced policy
$\bar\pi_{e,\theta}$.

\paragraph{Proof.}
From Equation~\ref{eq:app_score_policy},
\begin{equation}
\log
\frac{\bar\pi_{e,\theta}(y\mid x)}
     {\pi_0(y\mid x)}
=
u_{e,\theta}(x,y)
-
\log Z_{e,\theta}(x).
\label{eq:app_logratio_proof}
\end{equation}
Subtracting the expressions for $y^+$ and $y^-$ cancels the common
normalization term $\log Z_{e,\theta}(x)$ and yields
Equation~\ref{eq:app_dpo_form}. \hfill$\square$

\paragraph{Relation to regularized reward optimization.}
For fixed $\theta$, for any regularization coefficient $\eta>0$, define an implicit reward
\begin{equation}
R_{e,\theta}(x,y)
=
\eta\,u_{e,\theta}(x,y).
\label{eq:app_implicit_reward}
\end{equation}
For an arbitrary response distribution $\pi(\cdot\mid x)$, consider
\begin{equation}
J(\pi)
=
\mathbb E_{y\sim\pi(\cdot\mid x)}
[R_{e,\theta}(x,y)]
-
\eta
D_{\mathrm{KL}}
\!\left(
\pi(\cdot\mid x)
\Vert
\pi_0(\cdot\mid x)
\right).
\label{eq:app_regularized_reward}
\end{equation}
Substituting Equation~\ref{eq:app_score_policy} gives
\begin{equation}
J(\pi)
=
\eta\log Z_{e,\theta}(x)
-
\eta
D_{\mathrm{KL}}
\!\left(
\pi(\cdot\mid x)
\Vert
\bar\pi_{e,\theta}(\cdot\mid x)
\right),
\label{eq:app_reward_kl_identity}
\end{equation}
so $\bar\pi_{e,\theta}$ is the unique optimizer whenever the usual support
conditions hold. Because the implicit reward depends on $\theta$, this characterizes the induced policy for a fixed controller; it is not a fixed-reward optimization theorem for training CAST.

This result concerns the \emph{score-induced policy}. It does not imply that
$\bar\pi_{e,\theta}$ equals CAST's actual autoregressive policy for every
task-specific score.

If \(\pi_\theta\) and \(\pi_0\) are normalized on this same response space under the scoring protocol, and the response scores satisfy:

\begin{equation}
s_{e,\theta}(x,y)
=
\log\pi_\theta(y\mid x),
\qquad
s_{e,0}(x,y)
=
\log\pi_0(y\mid x),
\label{eq:app_full_logp_special}
\end{equation}
then $Z_{e,\theta}(x)=1$ and
$\bar\pi_{e,\theta}=\pi_\theta$ exactly.

For the length-normalized score
\begin{equation}
s_{e,\theta}(x,y)
=
\frac{1}{|y|}
\log\pi_\theta(y\mid x),
\label{eq:app_mean_logp_special}
\end{equation}
the induced policy remains well defined but generally differs from
$\pi_\theta$. The DPO-form equivalence in
Equation~\ref{eq:app_dpo_form} remains exact at the objective level.

\subsection{Local Hidden-State Controllability}
\label{app:local_controllability}

This section proves the single-site local-improvability statement used to
motivate DHSA.

Fix a structural component $i$. Let
$\Gamma_i\in\mathbb{R}^{d_i\times d_i}$ parameterize the shared local
intervention
\begin{equation}
h_{t,i}
\longmapsto
h_{t,i}+\Gamma_i h_{t,i},
\label{eq:app_local_intervention}
\end{equation}
and let $J_e(\Gamma_i)\in\mathbb{R}$ be a preference objective of the resulting
controlled model, with larger values preferred.

Define
\begin{equation}
G_{i,e}
=
\left.
\nabla_{\Gamma_i}
J_e(\Gamma_i)
\right|_{\Gamma_i=0}.
\label{eq:app_local_gradient}
\end{equation}

Assume $J_e$ is locally $L_i$-smooth in Frobenius norm around
$\Gamma_i=0$, i.e.,
\begin{equation}
\|
\nabla J_e(\Gamma)
-
\nabla J_e(\Gamma')
\|_F
\leq
L_i
\|
\Gamma-\Gamma'
\|_F
\label{eq:app_smoothness}
\end{equation}
throughout a Frobenius ball of radius $r_i>0$ around the origin.

\paragraph{Proposition 2 (single-site rank-one improvement).}
If $G_{i,e}\neq0$, then there exists a rank-one matrix
$D_i=u_i v_i^\top$ with $\|D_i\|_F=1$ and an $\epsilon_0>0$ such that
\begin{equation}
J_e(\epsilon D_i)
>
J_e(0)
\qquad
\text{for all }
0<\epsilon<\epsilon_0.
\label{eq:app_local_improve}
\end{equation}

More specifically, let
\begin{equation}
\gamma_i
=
\sigma_{\max}(G_{i,e})>0,
\label{eq:app_top_singular}
\end{equation}
and let $u_i,v_i$ be corresponding unit singular vectors. Then
$D_i=u_i v_i^\top$ satisfies
\begin{equation}
\langle
G_{i,e},
D_i
\rangle_F
=
\gamma_i.
\label{eq:app_svd_direction}
\end{equation}
By the standard smoothness lower bound~\citep{nocedal2006numerical},
\begin{equation}
J_e(\epsilon D_i)
-
J_e(0)
\geq
\epsilon\gamma_i
-
\frac{L_i}{2}\epsilon^2.
\label{eq:app_local_bound}
\end{equation}
Hence any
\begin{equation}
0<\epsilon<\min\!\left\{r_i,\frac{2\gamma_i}{L_i}\right\},
\label{eq:app_epsilon_range}
\end{equation}
strictly improves the local preference objective. \hfill$\square$

The proposition is an existence statement over the admissible local
state-control space. It does not guarantee that every interface satisfies
$G_{i,e}\neq0$, that an RCM-selected interface satisfies the premise, or that a
single interface can reproduce an arbitrary globally optimized policy.

\paragraph{Constant-vector special case.}
For an additive controller
$h_{t,i}\mapsto h_{t,i}+v_i$ with trainable
$v_i\in\mathbb{R}^{d_i}$, the same argument applies directly in vector space.
If
\begin{equation}
\left.
\nabla_{v_i}J_e(v_i)
\right|_{v_i=0}
\neq0,
\label{eq:app_vector_gradient}
\end{equation}
then a sufficiently small step in the normalized gradient direction strictly
improves $J_e$.

Thus both the constant and low-rank CAST actuator classes can contain local
improvement directions when their corresponding local gradients are nonzero.

\subsection{RCM Effects as Local-Sensitivity Proxies}
\label{app:rcm_proxy}

This section formalizes the weaker connection between finite RCM effects and
local controllability. It does not claim that RCM identifies an optimal control
site.

Let $\xi_i^a$ and $\xi_i^b$ denote the two registered internal quantities
compared by an RCM contrast. Depending on the contrast, $\xi_i$ can be a
component state or a residual write. Define the interpolation
\begin{equation}
\xi_i(\lambda)
=
(1-\lambda)\xi_i^a
+
\lambda\xi_i^b,
\qquad
\lambda\in[0,1],
\label{eq:app_rcm_path}
\end{equation}
and
\begin{equation}
\phi_{i,e,x}(\lambda)
=
C_e
\!\left(
M_{i\leftarrow\xi_i(\lambda)};
x
\right).
\label{eq:app_path_score}
\end{equation}

Assume $\phi_{i,e,x}$ is absolutely continuous on $[0,1]$ and differentiable
almost everywhere. Then the fundamental theorem of calculus gives
\begin{equation}
\Delta_{i,e}^{a\rightarrow b}(x)
=
\phi_{i,e,x}(1)-\phi_{i,e,x}(0)
=
\int_0^1
\phi'_{i,e,x}(\lambda)
\,d\lambda.
\label{eq:app_ftc}
\end{equation}
When $C_e$ is differentiable with respect to $\xi_i$,
\begin{equation}
\phi'_{i,e,x}(\lambda)
=
\left\langle
\nabla_{\xi_i}C_e,
\,
\xi_i^b-\xi_i^a
\right\rangle,
\label{eq:app_directional_sensitivity}
\end{equation}
where the gradient is evaluated at the intervened state
$\xi_i(\lambda)$. This is the same path-integral identity underlying
path-based attribution methods~\citep{sundararajan2017axiomatic}.

\paragraph{Proposition 3 (nonzero RCM implies path sensitivity).}
Under the conditions above,
\begin{equation}
\Delta_{i,e}^{a\rightarrow b}(x)\neq0
\quad\Longrightarrow\quad
\exists\lambda^\star\in(0,1)
\text{ such that }
\phi'_{i,e,x}(\lambda^\star)\neq0.
\label{eq:app_proxy_exist}
\end{equation}

\paragraph{Proof.}
If $\phi'_{i,e,x}(\lambda)=0$ almost everywhere on $[0,1]$, then
Equation~\ref{eq:app_ftc} gives
$\Delta_{i,e}^{a\rightarrow b}(x)=0$, contradicting the premise.
\hfill$\square$

If $\phi'$ is continuous, the mean-value theorem gives the stronger statement
\begin{equation}
\exists\lambda^\star\in(0,1):
\qquad
\phi'_{i,e,x}(\lambda^\star)
=
\Delta_{i,e}^{a\rightarrow b}(x),
\label{eq:app_mvt}
\end{equation}
because the path parameter has unit length.

\paragraph{Endpoint sensitivity.}
Suppose additionally that $\phi'$ is $L_{\mathrm{path}}$-Lipschitz in
$\lambda$:
\begin{equation}
|
\phi'(\lambda)-\phi'(\lambda')
|
\leq
L_{\mathrm{path}}
|\lambda-\lambda'|.
\label{eq:app_path_lipschitz}
\end{equation}
Then
\begin{equation}
\left|
\phi'(1)
-
\Delta_{i,e}^{a\rightarrow b}(x)
\right|
\leq
\frac{L_{\mathrm{path}}}{2}.
\label{eq:app_endpoint_bound}
\end{equation}
Therefore, if
\begin{equation}
\left|
\Delta_{i,e}^{a\rightarrow b}(x)
\right|
>
\frac{L_{\mathrm{path}}}{2},
\label{eq:app_endpoint_condition}
\end{equation}
the endpoint directional derivative is necessarily nonzero and has the same
sign as the finite RCM contrast.

This additional smoothness condition is stronger than that required for
Proposition~3 and is not assumed for all experiments.

\paragraph{Zero and Patch paths.}
For RCM-Zero,
\begin{equation}
\xi_i^a=0,
\qquad
\xi_i^b=z_i^{\mathrm{native}},
\label{eq:app_zero_path}
\end{equation}
so the contrast integrates sensitivity along the path from write removal to the
native residual write.

For RCM-Patch,
\begin{equation}
\xi_i^a=h_i^{\mathrm{ref}},
\qquad
\xi_i^b=h_i^{\mathrm{src}},
\label{eq:app_patch_path}
\end{equation}
so the contrast integrates sensitivity from the receiving state to the
specified source state. For a prototype patch, $h_i^{\mathrm{src}}$ is replaced
by $\bar h_{i,e}^{\mathrm{src}}$ from Equation~\ref{eq:app_patch_prototype}.
The reported Site experiments use this prototype form.

\paragraph{Stochastic generation protocols.}
The path-integral statement applies directly to deterministic differentiable
scores and, more generally, to an expected score when differentiation may be
exchanged with the expectation. A single realized discrete generation is not a
differentiable function of the intervention state. For such protocols, RCM
remains a well-defined finite intervention contrast, but
Equation~\ref{eq:app_ftc} is not invoked for that individual sampled
realization. This distinction is especially relevant when $C_e$ is evaluated
through sampled open generation followed by an external reward model.

\end{document}